\pdfoutput=1
\documentclass[manuscript,screen]{acmart}

\usepackage{amsmath}
\usepackage{booktabs}
\usepackage{algorithm} 
\usepackage{algorithmic} 
\usepackage{subfig}
\usepackage{graphicx}

\AtBeginDocument{%
  }

\setcopyright{acmcopyright}
\copyrightyear{2022}
\acmYear{2022}
\acmDOI{10.1145/nnnnnnn.nnnnnnn}

\begin{document}

\title{FedEgo: Privacy-preserving Personalized Federated Graph Learning with Ego-graphs}

\author{Taolin Zhang}
\email{zhangtlin3@mail2.sysu.edu.cn}
\author{Chuan Chen}
\email{chenchuan@mail.sysu.edu.cn}
\authornote{Corresponding author}
\author{Yaomin Chang}
\email{changym3@mail2.sysu.edu.cn}
\author{Lin Shu}
\email{shulin@mail2.sysu.edu.cn}
\author{Zibin Zheng}
\email{zhzibin@mail.sysu.edu.cn}
\affiliation{%
    \institution{Sun Yat-sen University}
    \streetaddress{School of Computer Science and Engineering, Higher Education Mega Center, Panyu District}
    \city{Guangzhou}
    \country{China}
}

\renewcommand{\shortauthors}{Taolin Zhang, Chuan Chen, Yao min Chang, Lin Shu and Zibin Zheng}

\begin{abstract}
    As special information carriers containing both structure and feature information, graphs are widely used in graph mining, e.g., Graph Neural Networks (GNNs). However, in some practical scenarios, graph data are stored separately in multiple distributed parties, which may not be directly shared due to conflicts of interest. Hence, federated graph neural networks are proposed to address such data silo problems while preserving the privacy of each party (or client). Nevertheless, different graph data distributions among various parties, which is known as the statistical heterogeneity, may degrade the performance of naive federated learning algorithms like FedAvg. In this paper, we propose FedEgo, a federated graph learning framework based on ego-graphs to tackle the challenges above, where each client will train their local models while also contributing to the training of a global model. FedEgo applies GraphSAGE over ego-graphs to make full use of the structure information and utilizes Mixup for privacy concerns. To deal with the statistical heterogeneity, we integrate personalization into learning and propose an adaptive mixing coefficient strategy that enables clients to achieve their optimal personalization. Extensive experimental results and in-depth analysis demonstrate the effectiveness of FedEgo.
\end{abstract}



\maketitle

\section{Introduction}
Graph Neural Networks (GNNs) have shown incredible performance in distilling information from graph data and deriving expressive node embedding that facilitates downstream tasks such as node classification and link prediction.
Nevertheless, previous GNN works focus on centralized node representation learning without taking into account the common existence of data silo problems in the real world. In a traditional data silo situation, the data are stored across several distributed parties and they're only allowed to be accessed privately. As a result, how to collaborate separate graphs from local data owners while preserving privacy for the training of a high-quality graph-based model is a crucial problem.

\begin{figure}[t]
    \centering
    \includegraphics[width=.8\textwidth]{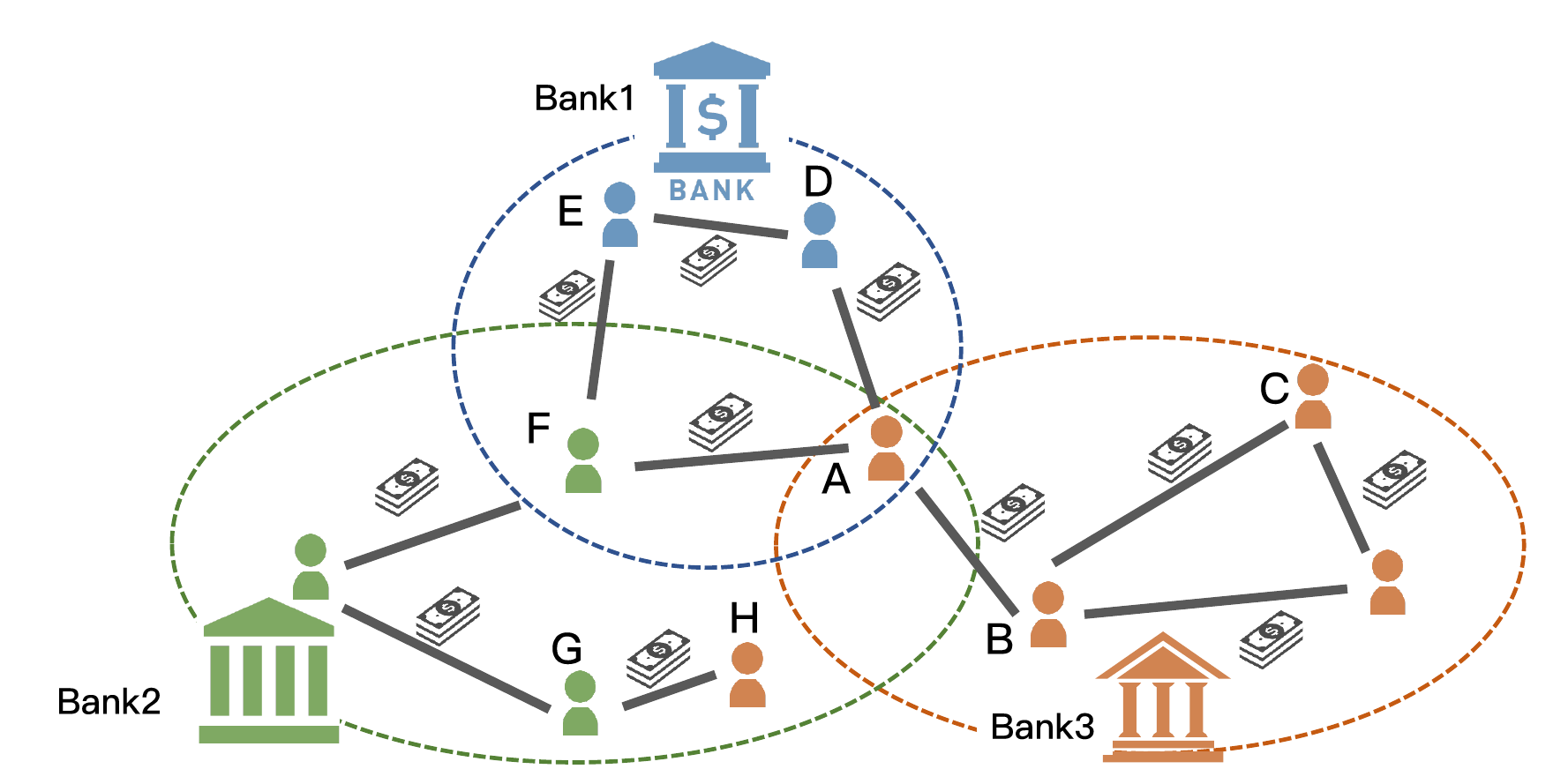}
    \caption{Motivating scenario in a financial system: Suppose there are three banks in the system, which are marked with different colors. In the system, a graph is formed with customers as nodes and their transactions as edges. People of the same color as the bank represent the bank's target customers who shares the same label such as wealthy. Customers in different colors hold different labels and banks desire to distinguish them, which forms a typical node classification task. The dashed lines indicate the detection range of the corresponding bank, some of which may overlap since someone is likely to use multiple banks. The bank has access to the relevant information about the nodes as well as the transactions between them in the detection range, and most of the nodes are its target customers just as the records of customers A, B, and C storing in Bank 3. However, some of its target customers may only make transfers in other banks without being detected such as customer H. In such realistic system, different target markets give rise to the statistical heterogeneity of the graph data among banks.}
    \label{motivation}
\end{figure}

Federated learning (FL), a technique that decouples the implementation of machine learning from the requirement for direct data sharing, has shown great promise in training models collaboratively while preserving data privacy  \cite{li2020federated}. The key idea of federated learning is to train a global model in a central server with the contribution of local data owners (or clients). When dealing with graph data, an intuitive idea is to combine naive federated algorithms with graph neural networks directly.
Nevertheless, naive FL algorithms based on weight aggregation such as FedAvg does not benefit from the structure information in the graph data, thereby may have poor performance in graph mining. For an organic combination of GNN and FL that we term as federated graph learning, the structure information of graph data is meant to be fully utilized.

Moreover, federated graph learning suffers from the highly non-independent identically distributed (non-IID) problem.
In the presence of the statistical heterogeneity among clients which commonly appears in the task of graph mining, a single global model might not generalize well on the local data of all clients  \cite{jiang2019improving}. As a consequence, it is necessary to integrate personalization into federated graph learning instead of training a single consensus model. That is to say, clients will adapt the global model to their own dataset and train their local models for personalization, which we term personalized federated graph learning.

Take a practical problem in the financial system as an example. There are many banks in a city constituting the financial system, as shown in Fig. \ref{motivation}. The records of each bank are separately stored without sharing with others because of privacy concerns and conflicts of interest. With customers as nodes and transactions as edges, a local graph could be derived from the records stored in a bank's database. Graph mining could be further applied and one of the most frequently applied tasks may be node classification such as distinguishing different kinds of customers. Furthermore, banks may also collaborate to improve the generalization ability of their local models. In many cases, however, various banks may target different markets and the distributed graph data are in a severe non-IID scenario, making the situation difficult. In other words, the local dataset of a bank can be regarded as a limited observation of the real world, and the key incentive for a bank to participate in collaboration is to reduce its local generalization error with the help of other banks' data. In light of these observations, a bank needs to achieve the trade-off between the benefit from the collaboration and the disadvantage brought by the potentially statistical heterogeneity, which forms a typical personalized federated graph learning.

In general, this realistic scenario poses three main challenges in personalized federated graph learning:
\begin{itemize}
    \item[$\bullet$] \textbf{Challenge 1:} How to make full use of the structure information of graph data during federated training? Since topological information is an indispensable part of graph mining, it is important to integrate them into federated learning organically.
    \item[$\bullet$] \textbf{Challenge 2:} How to mitigate the issue of the potentially non-IID graph data in the federated learning framework? Observations from different angles of the real dataset lead to severe non-IID graph data and prevent naive FL algorithms from performing well.
    \item[$\bullet$] \textbf{Challenge 3:} How to achieve an optimal trade-off between the benefit and the disadvantages of the collaboration? Under the potential non-IID scenario, the ideal situation for a client is to utilize others' data to compensate for its local dataset while minimizing the harm induced by the statistical heterogeneity among each other.
\end{itemize}

The challenges above motivate us to design FedEgo, a personalized federated graph learning system based on ego-graphs. FedEgo is capable of handling the challenges above and preserves privacy by keeping data anonymous in terms of structure and feature.
\begin{itemize}
    \item[$\bullet$] \textbf{To address challenge 1}, it is feasible to view the graph as a family of $k$-hop ego-graphs with structures and node features. Ego-graph, a sampled subgraph with up to k-hop neighbors of the center node, is a kind of useful information carrier and enables message passing of structure and feature information during the graph mining.
        FedEgo extracts topological information from ego-graphs by applying GraphSAGE  \cite{hamilton2017inductive} over them. Furthermore, only the local topological information can be derived from the sampled ego-graphs and it is impossible to restore the original structure, which means that they are structure-anonymous.

    \item[$\bullet$] \textbf{To address challenge 2}, we train a global model in the central server to handle the non-IID graph data. Taking consideration of challenge 1, we seek to develop a global model which is capable of capturing the structural information that ego-graphs contains.
        To this end, inspired by  \cite{arivazhagan2019federated,collins2021exploiting}, we introduce a network that composes of reduction layers and personalization layers to exploit shared low-dimensional embeddings and perform personalized graph mining, respectively. In FedEgo, the clients' model consists of both reduction layers and personalization layers for local training while the global model in the server consists of only the personalization layers for information distillation of the ego-graphs in the global dataset.
        By applying Mixup \cite{yoon2021FedMix,zhang2017mixup} within each batch in clients, mashed ego-graphs are generated, which are later sent to the server and form the global dataset.
        In the server, the ability of the global model to deal with non-IID graph data is developed by training over the uploaded mashed ego-graphs.
        The mashed ego-graphs, which contain only the mashed embedding and local structure, prevent the transmission of the raw data and protect privacy, and thus they are further feature-anonymous.

    \item[$\bullet$]  \textbf{To address challenge 3}, clients perform updates with the help of the global model.
        Clients will follow the vanilla algorithm of FedAvg   \cite{mcmahan2017communication} in reduction layers to reach a consensus in encrypting the ego-graphs, and then mix the local and global weights in personalization layers according to an adaptive mixing coefficient. The mixing coefficient, which contributes to achieving better personalization for each client, is adaptively determined by the the difference between the distribution of local and global datasets.
\end{itemize}
We validate FedEgo on the real-world datasets in non-IID settings to better simulate the application scenarios. Experimental results show that FedEgo significantly outperforms baselines,
which fully verifies the effectiveness of FedEgo. We summarized our main contribution as follows.
\begin{itemize}
    \item[$\bullet$] We introduce a novel personalized federated graph learning framework based on ego-graphs. FedEgo makes full use of structure information of graph data by applying GraphSAGE over ego-graphs.
    \item[$\bullet$] We apply Mixup over ego-graphs for privacy concerns and develop the global model's ability to capture structural information and deal with the non-IID graph data by training over the uploaded mashed ego-graphs.
    \item[$\bullet$] We design a strategy to adaptively learn a personalized model for a trade-off between the benefit and disadvantage of the collaboration.
    \item[$\bullet$] We conduct extensive experiments on widely used datasets and empirically demonstrate the superior performance of FedEgo under non-IID scenarios.
\end{itemize}

\section{Related Work}\label{RW}

\subsection{Federated Learning on Graphs}
Recently, federated learning on graphs have raised great interest and several federated graph frameworks have been proposed by leveraging the power of
federated learning and graph neural networks  \cite{chen2021fedgl,zhang2021subgraph,he2021fedgraphnn,wang2020graphfl,xie2021federated,pei2021decentralized}. GraphFL  \cite{wang2020graphfl} is a
model-agnostic meta learning approach designed for few-shot learning and D-FedGNN \cite{pei2021decentralized} is a distributed federated graph framework which allows collaboration among clients without a centralized server. FedSage+  \cite{zhang2021subgraph}, which trains a missing neighbor generator to recover the missing edges cross clients, mainly targets distributed subgraph systems that are not very common in practice. FedGL \cite{chen2021fedgl} uploads the prediction results and embeddings for global information of nodes and FedGCN \cite{yao2022fedgcn} exchanges average information about the node's neighbors among clients. Both of them suffer from severe privacy problems since others know whether a specific node is in a certain client's local dataset. FedEgo, by contrast, preserves privacy for node classification tasks in a realistic setting by keeping the data anonymous in terms of structure and feature.

\subsection{Personalized Federated learning}
Personalization in federated learning has attracted much attention and has been widely explored in recent approaches  \cite{arivazhagan2019federated,jiang2019improving,deng2020adaptive,hanzely2020federated,mansour2020three}.   \cite{arivazhagan2019federated} proposes a neural network with base layers for federated averaging and personalization layers for personalization.   \cite{collins2021exploiting} is a further extension of  \cite{arivazhagan2019federated} and obtains low-dimensional embedding which accelerates the convergence speed of the training process. Some other works   \cite{jiang2019improving,chen2018federated} borrow ideas from MAML and   \cite{jiang2019improving} considers federated learning as an instance of MAML\@. Moreover,  \cite{wang2020graphfl} introduces MAML into federated graph learning framework and designs GraphFl for few-shot learning.
Furthermore, it is proved effective to improve the performance of the personalized model by using a mixture of global model and local model  \cite{deng2020adaptive,mansour2020three}, multi-task learning  \cite{smith2017federated} and adding proximal terms to perform local fine-tuning  \cite{hanzely2020federated}. Different from existing work, herein we discuss how to achieve better personalization based on the difference of local and global distributions with theoretical justification.

\begin{figure*}[t]
    \centering
    \subfloat[]{
        \includegraphics[width=.25\textwidth]{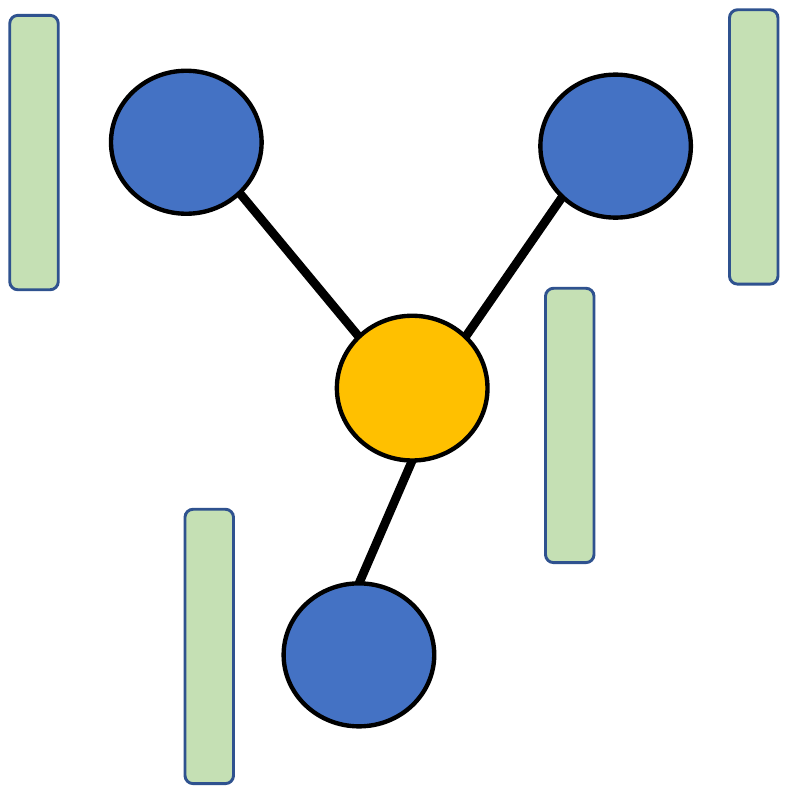}
        \label{ego_graph}
    }
    \subfloat[]{
        \includegraphics[width=.45\textwidth]{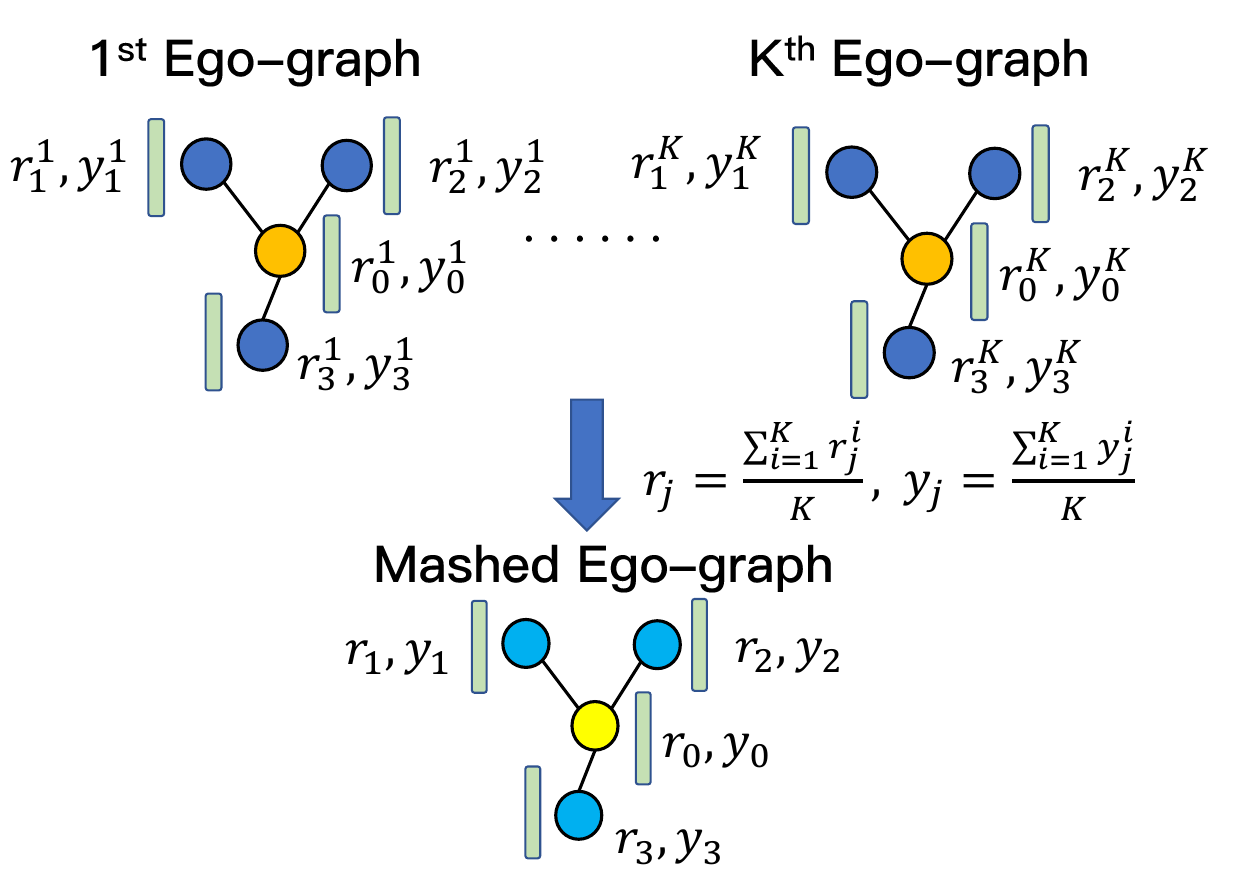}
        \label{Mixup}
    }
    \caption{(a) Illustration of 1 hop ego-graph.
        (b) Illustration of the alignment and Mixup among a batch of ego-graphs. The center nodes are aligned together and their neighbors are extended recursively. The reduction embedding $r$ and one-hot label $y$ are averaged according to the alignment. }
\end{figure*}

\section{Preliminaries}\label{Pre}
\subsection{Ego-Graph}\label{definition_ego_graph}
Let $G = (V, E)$ be a graph with $N$ nodes, where $V=\{v_1,\dots,v_N\}$ denotes the node set and $E \subseteq V \times V$ denotes the edge set.
Following  \cite{bai2016fast,zhu2020transfer}, we have the definition of a $k$-hop ego-graph.
\begin{definition}{($k$-hop ego-graph).}
    A graph $g_v=\{V_v,E_v\}$ is called a $k$-hop ego-graph centered at node $v$ if it has a $k$-layer centroid expansion such that the greatest shortest path rooted from $v_{i}$ has length $k$, i.e., $k=\max _{v_{i} \in V}|S(v, v_{i})|$, where $S(v, v_{i})$ is the shortest path from $v$ to $v_{i}$ and $|\cdot|$ denotes the length of the path.
\end{definition}

Given $k$, a depth-based representation of the graph can be described as a family of ego-graphs $G=\{g_{v_i}\}_{i=1}^{N}$  \cite{bai2016fast,zhu2020transfer}. The label of each node is retained in the ego-graph, which is represented as a one-hot vector.
As useful information carriers, ego-graphs will be sampled in each client for the training in FedEgo. In practice, we follow the method in GraphSAGE  \cite{hamilton2017inductive} and sample a fixed-size set of neighbors in each layer for the nodes.
Note that the sampled ego-graphs will be in a fixed shape and can be trained independently of the original graph.

When training with the family of ego-graphs, we do not concern about the concrete information in the original graph but only the local property. In this sense, there is no way of knowing where the center node is located in the original graph and one could not restore the original graph through the sampled ego-graphs. Therefore, the sampled ego-graphs are structure-anonymous.


\subsection{GraphSAGE}\label{GCN}
GraphSAGE  \cite{hamilton2017inductive} is a method applicable to inductive graph learning, with which we could enable the information flow in ego-graphs. It focus on the local property of the target node and aggregates feature from the neighborhood to obtain embedding for downstream tasks. Let $h_v^{(l)}$ denotes the embedding of node $v$ in the layer $l$ and the convolution layer with a mean aggregator can be defined as
\begin{equation}\label{loss}
    h_v^{(l)}=\sigma(W^{(l)}\cdot h_v^{(l-1)}+W^{(l)}\cdot Mean(\{h_u^{(l-1)},u\in N(v)\})),
\end{equation}
with $W^{(l)}$ as the weight matrix and $N(v)$ as the 1-hop neighbors of $v$.
In paricular, $h_v^{(0)}$ denotes the input features.
For supervised node classification tasks, we use cross-entropy as the loss function.
Let $(x,y)$ denotes the node samples with feature and label, $f_c:X \to S$ be the map function from feature space $X$ to the label space $S$, $\mathbf{1}_{y=c}$ be the indicator function of label $c$,  $W$ be the weight in the whole model. Then the cross-entropy loss could be formulated as
\begin{equation}\label{loss}
    \ell(W)=\mathbb{E}_{{x}, y \sim P}[\sum_{c=1}^{C} \mathbf{1}_{y=c} \log f_{c}({x}, {W})]=\sum_{c=1}^{C} P(y=c) \mathbb{E}_{{x} \mid y=c}[\log f_{c}({x}, {W})],
\end{equation}
where $P$ represents the probability vector of data distribution and $\sum_{c=1}^{C} P(y=c)=1$.
The distribution vector $P$ can be further obtained by following formula.
\begin{equation}\label{p}
    P=\frac{\sum y_{one-hot}}{|N|},
\end{equation}
where $y_{one-hot}$ is the one-hot vector of label and $|N|$ is the number of nodes.

\subsection{Federated Learning}
In federated learning, the main objective is to learn a global model by clients' collaboration and a typical implementation is FedAvg \cite{mcmahan2017communication}.
In FedAvg, clients perform local updates and global averages iteratively and finally learn a global model.
Formally, we assume that there are $N$ clients with $W_i$ as the local weight of client $i$. In a training epoch, the server first broadcasts the latest global weight $W_g$ to all the clients. Subsequently, clients load the weight $W_i=W_g$ and then perform local updates several times:
\begin{equation}
    W_i = W_i - \alpha \nabla L_i(W_i).
\end{equation}
After that, the server averages the uploaded local parameters and obtain the new global model:
\begin{equation}
    W_g = \frac{1}{N}\sum_{i=1}^{N} W_i.
\end{equation}
The process is repeated for convergence and a global model is learned for prediction.
FedAvg aggregates information from models whereas it ignores the difference between clients. Previous works show that it may perform poorly when it comes to severe non-IID situation \cite{xie2021federated}.

\subsection{Mixup}
Mixup  \cite{zhang2017mixup} is a data augmentation technique that apply linear combination over data samples to generate additional data.
In the federated framework, Yoon et al.  \cite{yoon2021FedMix} propose FedMix and apply Mixup over samples within each batch. Formally, given a batch of $n$ data samples $\{(x_i,y_i)\}_{i=1}^n$, FedMix constructs augmented data sample by averaging features and labels:
\begin{equation}\label{mixup}
    \left\{
    \begin{aligned}
        \tilde{x}=\frac{\sum_{i=1}^n x_i}{n} \\
        \tilde{y}=\frac{\sum_{i=1}^n y_i}{n}
    \end{aligned}
    \right.
\end{equation}

\begin{figure}[t]
    \centering
    \includegraphics[width=\textwidth]{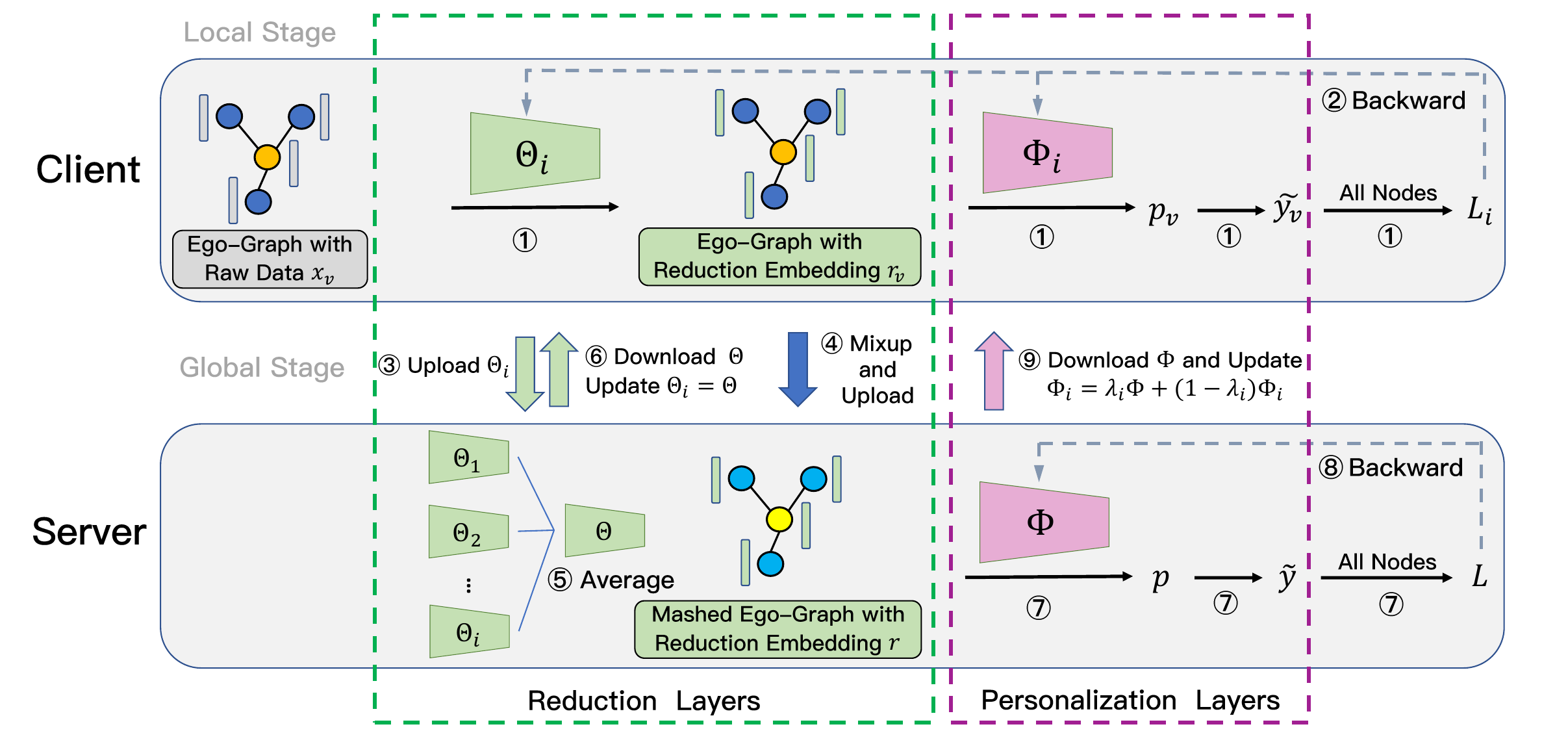}
    \caption{The detailed framework of FedEgo.}
    \label{model}

\end{figure}

\section{Proposed method: FedEgo}
In FedEgo, we expect the server to be capable of capturing both structural and feature information of non-IID graph data from clients. And then with the help of the global model, clients update their local model for better personalization.
To this end, there are different network architectures between clients and the server. Specifically, a client's local model consists of reduction layers and personalization layers, whereas there are only personalized layers in the server. Reduction layers are designed to exploit shared low-dimensional embeddings among clients while personalization layers are used for personalized graph mining.
To extend GraphSAGE over the ego-graphs, we follow \cite{hamilton2017inductive} and
sample a fixed-size set of neighbors for each given node. And then the training of FedEgo can be mainly divided into local stage and global stage.
(1) In the \textbf{Local Stage}, clients feed their local ego-graphs into reduction layers and obtain the low-dimensional embedding. Subsequently, Mixup is applied over ego-graphs to generate mashed ego-graph within each batch. Then the embedding obtained from reduction layers will be further fed into personalization layers. Eventually, each client calculates the loss and updates the parameters in reduction layers and personalization layers.
(2) In the \textbf{Global Stage}, clients upload parameters in reduction layers and the mashed ego-graphs to the server for collaboration. The server aggregates the parameters by applying FedAvg algorithm and updates the global personalization layers by training over the mashed ego-graphs. After that, all the parameters are sent back to clients.
Clients will then load the reduction layers and update their personalization layers by mixing the local and global weight.
The framework of FedEgo is illustrated in Fig. \ref{model} and Algorithm \ref{algorithm} demonstrates the training process.

\begin{algorithm}[htbp]
    \caption{The algorithm of FedEgo}
    \label{algorithm}
    \begin{algorithmic}[1]
        \REQUIRE Clients number $N$, node feature matrix $\big\{X_i\big\}$, initialized weight in reduction layers and personalization layers of each client $\Theta_i$ and $\Phi_i$, initialized weight in personalization layers of the global model $\Phi_g$
        \ENSURE Model parameters for each client $\big\{\Theta_i\big\}$ and $\big\{\Phi_i\big\}$
        \STATE Clients sample $k$ hop ego-graphs in the fixed shape from their local datasets and calculate local distribution vector $P_i$ by Eq. (\ref{p}).
        \STATE \textbf{while} not converge \textbf{do}
        \STATE \qquad \textbf{Local Stage:}
        \STATE \qquad \textbf{for} client $i = 1$ to $N$ \textbf{do in parallel}
        \STATE \qquad \qquad \textbf{for} each epoch \textbf{do}
        \STATE \qquad \qquad \qquad \textbf{for} each batch \textbf{do}
        \STATE \qquad \qquad \qquad \qquad \textbf{for} each node $v$ within the batch with feature $x_v\in X_i$ \textbf{do}
        \STATE \qquad \qquad \qquad \qquad \qquad $r_{v}=\Theta_i(x_v)$ // \textit{Local reduction layers}
        \STATE \qquad \qquad \qquad \qquad \qquad  $p_{v}=\Phi_i(r_{v})$ // \textit{Local personalization layers}
        \STATE \qquad \qquad \qquad \qquad \textbf{end}
        \STATE \qquad \qquad \qquad \qquad // \textit{Generate a mashed ego-graph}
        \STATE \qquad \qquad \qquad \qquad Calculate the embedding and the label of each node in the mashed ego-graph by Eq. (\ref{mashed})
        \STATE \qquad \qquad \qquad \qquad // \textit{Parameters update}
        \STATE \qquad \qquad \qquad \qquad Calculate loss $\ell_i$ by Eq. (\ref{loss})
        \STATE \qquad \qquad \qquad \qquad Set $\Theta_i \leftarrow \Theta_i-\eta \nabla \ell_i$, $\Phi_i \leftarrow \Phi_i-\eta \nabla \ell_i$,
        \STATE \qquad \qquad \qquad \textbf{end}
        \STATE \qquad \qquad \textbf{end}
        \STATE \qquad \textbf{end}

        \STATE \qquad \textbf{Global Stage:}
        \STATE \qquad // \textit{Averaging in reduction layers}
        \STATE \qquad Clients send local weight in reduction layers $\Theta_i$ and Server averages the parameters by $\Theta_g=\frac{\sum_{i=1}^{N}\Theta_i}{N}$
        \STATE \qquad // \textit{Training of the global model}
        \STATE \qquad Clients send all the mashed ego-graphs with embedding and label to Server
        \STATE \qquad \textbf{for} each epoch \textbf{do}
        \STATE \qquad \qquad \textbf{for} each batch \textbf{do}
        \STATE \qquad \qquad \qquad \textbf{for} each node with mashed embedding $r$ \textbf{do}
        \STATE \qquad \qquad \qquad \qquad  $p=\Phi_g(r)$ // \textit{Global personalization layers}
        \STATE \qquad \qquad \qquad \textbf{end}
        \STATE \qquad \qquad \qquad // \textit{Parameters update}
        \STATE \qquad \qquad \qquad Calculate loss $\ell_g$ by Eq. (\ref{loss})
        \STATE \qquad \qquad \qquad Set $\Phi_g \leftarrow \Phi_g-\eta \nabla \ell_g$
        \STATE \qquad \qquad \textbf{end}
        \STATE \qquad \textbf{end}
        \STATE \qquad // \textit{Parameters update for clients}
        \STATE \qquad Server calculates the gloal distribution vector $P_g$ by Eq. (\ref{p})
        \STATE \qquad \textbf{for} client $i = 1$ to $N$ \textbf{do in parallel}
        \STATE \qquad \qquad Calculate $EMD_i$ and $\lambda_i$ by Eq. (\ref{EMD}) and (\ref{lambda})
        \STATE \qquad \qquad Set $\Theta_i\leftarrow \Theta_g$, $\Phi_i\leftarrow \lambda_i\Phi_g+(1-\lambda_i)\Phi_i$
        \STATE \qquad \textbf{end}
        \STATE \textbf{end}

    \end{algorithmic}
    \label{Algorithm}
\end{algorithm}

\subsection{Local Stage}

\subsubsection{Reduction Layers: Multi-Layer Perception}
In the federated framework, data distribution between clients is in a severe non-IID situation, which gives rise to the difficulty to train a central model. However, these heterogeneous data may share a common representation despite having various labels  \cite{collins2021exploiting} and hence reduction layers are proposed to capture the common low-dimensional embeddings across clients, as shown in the green part of Fig. \ref{model}.
And the embedding obtained by reduction layers, which we term reduction embedding, may facilitate subsequent calculations while also protecting data privacy due to the reduction of dimensionality.
Besides, it has been observed that deeply stacking the layers often results in significantly worse performance for GNNs, which is called over-smoothing \cite{cai2020note}.
To avoid such a problem, herein we dig out the reduction embedding using a multi-layer perception since 2-layer GNN will be applied in personalization layers for graph mining.
Formally, consider $x_v\in\mathbb{R}^d$ as the $d$-dimension raw feature of node $v$ and a $l_r$ layers perception, the hidden embedding can be calculated as:
\begin{equation}\label{base}
    r_{v}^{(l)}=\sigma\big(W^{(l)}_{r}\cdot r_{v}^{(l-1)}+b^{(l)}_{r}\big),
\end{equation}
where $W^{(l)}_{r}$ and $b^{(l)}_{r}$ is the weight and the bias of layer $l$ respectively, $\sigma$ is the activation function and $r_{v}^{(l)}$ represents the hidden embedding of node $v$ obtained by layer $l$. Specifically, we take the raw feature as input of the reduction layers and we have $r_v^{(0)}=x_v$. The final embedding $r_v^{(l_r)}$ is regarded as the reduction embedding. For convenience, we term the whole reduction layers as $\Theta$.

\subsubsection{Mashed Ego-graphs: Mixup within Each Batch}
In FedEgo, clients will generate mashed ego-graphs for the training of the global model and further send them to the server rather than raw data. All mashed ego-graphs uploaded are structure-anonymous, which means that the server has no way of knowing the structure of the original graphs and whether a specific node exists in the local dataset of a certain client.
The mashed ego-graphs are in the same shape as the ego-graphs and can be obtained by mixing up the reduction embedding of ego-graphs within each batch. As a popular data augmentation technique for traditional data, Mixup, however, is not directly applicable to graph data because of the alignment problem. The feature matrices of traditional data samples share the same dimension and can be used for element-wise operations while the irregularity of structures in graph data limits the element-wise calculation and various ways to mix up may give rise to different impacts on the training.

Mixup, on the other hand, is relatively easy to be applied over ego-graphs in the fixed shape. The number of $k$-hop neighbors in such ego-graphs is only determined by $k$ and the size of the neighbors set $n$. For convenience, we sort the nodes based on its layer to assign each node a specific position and obtain the position sets $Q_{i}$ in the $i$-th layers, i.e., ${Q_0}=\{0\}$, ${Q_1}=\{1, 2,\dots n\}$, ${Q_2}=\{n+1, n+2, \dots n^2\}$ and so on. The node in a specific position connects to its corresponding neighbors in the next layer. For example, the neighbors of node $1$ in the second layer can be represented as $\{n+1,n+2,\dots 2n\}$. The position within the same layer can be assigned arbitrarily as long as the special connectivity between layers is maintained.
To apply Mixup over ego-graphs, we could first align the center nodes, i.e., the node $0$ of all the ego-graphs within a batch together and then recursively extend their neighbors to assign each node a specific position in its corresponding layer. Then the embedding of each position in the mashed ego-graph can be obtained by simply averaging the embedding of the corresponding nodes according to the alignment, which is illustrated in Fig \ref{Mixup}.
Formally, given the embedding and the one-hot label of the $j$-th position in the $i$-th ego-graphs $r_j^i$ and $y_j^i$ according to the alignment, we obtain the mashed embedding $r_j$ as well as the label $y_j$ of the nodes in the mashed ego-graphs.
\begin{equation}\label{mashed}
    r_j=\frac{\sum_{i=1}^n r_j^i}{n}
    , y_j=\frac{\sum_{i=1}^n y_j^i}{n}
\end{equation}
The mashed ego-graphs average the reduction embedding to prevent privacy leakage, making them feature-anonymous in addition to structure-anonymous.
Besides, there are lots of ways to align ego-graphs though, all of them have the same effect on a specific GNN model according to Theorem \ref{theorem}.

\begin{theorem}\label{theorem}
    Mashed ego-graphs generated from all sorts of alignments are equivalent for the training
    of linear GraphSAGE layers without activation functions. If there are two different
    alignments $A$ and $B$, the final embedding of the center node after aggregation
    under alignment $A$ and $B$ will be the same.
\end{theorem}
The detailed proof of Theorem \ref{theorem} is provided in Appendix. As Theorem \ref{theorem} shows, linear GraphSAGE layers avoid the impact that different alignments bring, and hence they could be used as the personalization layers without incuring bias.
However, the unbiased estimate of the embedding may not always guarantee better performance when dealing with complex scenarios, such as the distribution of highly non-IID data among clients.
Due to the high statistical heterogeneity, activation function may be preferable in the personalization layers since it empowers the model's ability to learn complex patterns while also introduces bias that will alleviate the severe non-IID issues. In developments test we found a slight improvement of GraphSAGE with activation over the linear one and thus focus on the former for the rest of our experiments.

\subsubsection{Personalization Layers: GraphSAGE over Ego-graphs and Classification}
Once obtained the reduction embedding, we further feed the ego-graph into linear GraphSAGE layers to forward implement graph mining, as can be seen in the pink part of Fig. \ref{model}. Formally, we have the personalization layers
\begin{equation}\label{deepLayer}
    p_{v}^{(l)}=\left(W_{p}^{(l)}\cdot p_v^{(l-1)}+W_{p}^{(l)}\cdot \frac{\sum_{u\in N(v)}{p_{u}^{(l-1)}}}{|N(v)|}\right),
\end{equation}
where $p_v^{(l)}$ represents the hidden embedding of node $v$ obtained by layer $l$. And we have $p_v^{(0)}=r_v^{(l_r)}$, i.e., using the reduction embedding as input.
Moreover, it is worth noting that we use linear GraphSAGE layers without activation functions between them.
After that, predictions are generated by a subsequent linear classifier with a softmax function, and the loss is calculated eventually.
For convenience, we term the whole personalization layer as $\Phi$.

\subsection{Global Stage}

\subsubsection{Global training}
In the global stage, the server trains a global model over the mashed ego-graphs uploaded by clients and updates the parameters in the global model.
There exist the same personalization layers in the server as in clients and we have GNN layers defined like Eq. (\ref{deepLayer}) with a classifier in the global model.

\begin{equation}\label{globalDeepLayer}
    p_{v}^{(l)}=\sigma \left(W_{g}^{(l)}\cdot p_u^{(l-1)}+W_{g}^{(l)}\cdot \frac{\sum_{u\in N(v)}{p_{u}^{(l-1)}}}{|N(v)|}\right),
\end{equation}
where $W_{g}^{(l)}$ denotes the $l$-th layer global personalization layers weight.

When considering the commnication cost, another variant of FedEgo is to replace the reduction layers with GNN and the personalization layers with MLP.
An incentive in this variant is that clients simply need to upload their reduction embedding of the center nodes instead of the structure in the mashed ego-graphs.
However, the information flow of topological structure is implicit in the form of the reduction embedding, making it rather challenging for the server to train especially in the case of highly non-IID data. Averaging the GNN parameters in the reduction layers may also be inappropriate since different data distribution among clients contributes to various node connection patterns.
Consequently, clients will receive limited or even negative help because of the poor collaboration.

When using the GNN layers as the personalization layers, the generalization ability of the global model is obviously better than any local one because of the access to the encrypted data among all clients.
After that, clients perform the following updates to enhance the performance of their local models.

\subsubsection{Averaging in Reduction Layers}
Aiming at exploiting shared representations among clients,
the parameters in reduction layers are updated by coordinate-wise weight averaging.
Formally, we have the following update in reduction layers.

\begin{equation}\label{base_avg}
    \Theta_{g} = \frac{1}{N}\sum_{i=1}^{N} \Theta_{i},
\end{equation}

\begin{equation}\label{base_update}
    \Theta_{i} = \Theta_{g},
\end{equation}
where $\Theta_{i}$ represents the reduction layers in client $i$ and $\Theta_{g}$ denotes avgeraged one.
With the update in reduction layers, clients reach a consensus to some degree even if graph data from their datasets are potentially non-IID. In other words, clients encrypt their raw data in the same way and project their data into a low-dimensional space.

\subsubsection{Mixing in Personalization Layers}
When updating parameters in personalization layers, clients will mix local and global weight to achieve the trade-off of advantages and drawbacks of the global model. Given a mixing coefficient $\lambda_i$ for client $i$, we update the personalization layers as follows.
\begin{equation}\label{deep_update}
    \Phi_{i} = \lambda_i \cdot \Phi_{g}+(1-\lambda_i)\cdot \Phi_{i},
\end{equation}
where $\Phi_{i}$ represents the personalization layers in client $i$ and $\Phi_{g}$ denotes global one.
Intuitively, the strategy to select mixing coefficient hinges on the diversity between local and global distributions. With great diversity, the mixing coefficient is expected to be large in that the client tends to gain more from the global model to correct its deviations and reduce the local generalization error. Otherwise, $\lambda$ needs to be small for better personalization of the local model.


\subsubsection{Adaptive Mixing Coefficient for Each Client}
It is obvious that a fixed $\lambda$ is inappropriate for all the clients due to the potential statistical heterogeneity.
We further propose an adaptive strategy to select $\lambda$ with theoretical analysis, thereby adjusting $\lambda$ for each client to achieve better personalization.
For convenience, we termed the model weights in clients and the server as local and global weight, respectively. First, following  \cite{zhao2018federated}, a metric is designed in Definition \ref{definition_wd} to measure the distance between local and global weights. We only analyze the personalization layers here since the reduction layers are frozen by averaging. Then we bound the weight divergence in Theorem \ref{theorem2}.
\begin{definition}\label{definition_wd}
    Let $\Phi_i$ denotes the local weight in personalization layers on client $i$, $\Phi_g$ denotes the global one, then
    the weight divergence $WD_i$ is defined as the distance between local and global weight: $WD_i=\|\frac{\Phi_i-\Phi_g}{\Phi_g}\|$.
\end{definition}

\begin{theorem}\label{theorem2}
    Given $N$ clients with $K$ samples of nodes $(x_k,y_k)_{k=1}^{K}$ drawn i.i.d from local data distribution $P_{i}$ for client $i \in[N]$ and the global data distribution $P_g$. Let train epochs for clients and server be the same $E_{c}=E_{s}=t-1$. Consider the update in personalization layers as a separate step and it is conducted every $t$ steps. The local weight of client $i$ and the global weight in the $t$-th step of the $T$-th epoch are denoted as $\Phi_{i,t}^{(T)}$ and $\Phi_{g,t}^{(T)}$, respectively. Let $\nabla_{{w}} \mathbb{E}_{{x} \mid y=c} \log f_{c}({x}, {w})$ is
    $L_{{x} \mid y=c}$-Lipschitz for each class $c \in[C]$ and the mixing coefficient $\lambda_i$, then the bound of weight divergence after $T$-th update is formulated as follows.

    \begin{equation}
        \begin{aligned}
            \|\Phi_{i,t}^{(T)}-\Phi_{g,t}^{(T)}\|
            \leq & (1-\lambda_i)a^{t-1}\|\Phi_{i,t}^{(T-1)}-\Phi_{g,t}^{(T-1)}\|                                                  \\
                 & +\eta(1-\lambda_i) \sum_{c=1}^{C}\|P_i(y=c)-P_g(y=c)\|(\sum_{j=0}^{t-2}a^{j} g_{\max }(\Phi_{g,t-2-j}^{(T)})),
        \end{aligned}
        \nonumber
    \end{equation}
    where $g_{\max }({w})=\max_{c=1}^{C}\|\nabla_{{\Phi}} \mathbb{E}_{{x} \mid y=c} \log f_{c}({x}, {w})\| \text { and } a=1+\eta \max _{c=1}^{C}L_{{x} \mid y=c}$.
\end{theorem}
The proof is provided in Appendix \ref{proof2}.
The second part on the right side includes the difference between the distribution of local and global datasets that is mainly reflected in $\sum_{c=1}^{C}\|P_i(y=c)-P_g(y=c)\|$, which is termed as earth mover distance ($EMD$). EMD is correlated with $\lambda_i$ and thus we can fine-tune $\lambda_i$ to affect the impact of EMD and optimize the weight divergence.
When the difference of local and global distribution is large (larger $EMD$), a larger $\lambda_i$ should be selected to pull the client closer to the global model. Otherwise, $\lambda_i$ should be small to fully achieve the personalization.
Both the local distribution vector $P_i$ and the global one $P_g$ can be calculated by Eq. (\ref{p}) even though there are only averaged data in the server. Then we can obtain $EMD_i$ of client $i$.

\begin{equation}\label{EMD}
    EMD_i=\sum_{c=1}^{C}\|P_i(y=c)-P_g(y=c)\|
\end{equation}

We further provide a formula to adaptively select $\lambda$ by introducing a hyperparameter $\gamma$ in Eq. (\ref{lambda}). It ensures $1-\lambda_i$ to be negative correlated with $EMD_i$ and $\lambda_i$ varies from 0 to 1 with $EMD_i$ increases within its range.
\begin{equation}\label{lambda}
    \lambda_i=(\frac{EMD_i}{2})^{\gamma}
\end{equation}

\section{Experiment}

\subsection{Datasets and experimental settings}
We conduct our experiments on four real datasets: Cora \cite{sen2008collective}, Citeseer \cite{sen2008collective}, CoraFull \cite{bojchevski2017deep}, and Wiki \cite{mernyei2020wiki}.
Table \ref{setting} shows the details of the datasets and the relevant settings.
In our experiments, we construct a local dataset for each client and a global dataset for final test, which are used for verifying the personalization and the generalization ability of the model, respectively.
We first sample nodes for global testing and delete them from the original dataset, leaving the remaining nodes for clients' local dataset.
To construct a label distribution skew scenario, the nodes are divided into different sets depending on their labels. Then each client randomly selects 3 labels as its major node labels and sample nodes from the corresponding sets to compose $80\%$ nodes in its local dataset. Random unselected nodes will be added as the remaining $20\%$ nodes.
In each client, 300 nodes are sampled for testing and 20\% nodes for validating, leaving other nodes for training. We choose 2 hop neighbors for each node and set the number of neighbors to be 6 when sampling ego-graphs.

\begin{table*}[htbp]
    \caption{The statistics and relevant setting of four datasets. |V| and |E| shows the number of node and the edges, respectively. \#C denotes the number of classes. $N$ indicates the number of clients chosen in the experiments. $\alpha_{global}$ and $\alpha_{local}$ are the sample rate of the global test dataset and local dataset, respectively. $lr$ is the learning rate of the Adam optimizer.}
    \centering
    \begin{tabular}{lccccccc}
        \toprule
        Dataset  & |V|   & |E|   & \#C & N  & $\alpha_{global}$ & $\alpha_{local}$ & $lr$   \\
        \midrule
        Cora     & 2708  & 5429  & 7   & 5  & 0.3               & 0.3              & 0.01   \\
        Citeseer & 3312  & 4715  & 6   & 5  & 0.3               & 0.3              & 0.01   \\
        Wiki     & 17716 & 52867 & 4   & 10 & 0.3               & 0.2              & 0.0003 \\
        CoraFull & 19793 & 63421 & 70  & 10 & 0.3               & 0.3              & 0.01   \\
        \bottomrule
    \end{tabular}
    \label{setting}
\end{table*}

\subsection{Comparison Methods}

We compare FedEgo with the following methods:
\begin{itemize}
    \item[$\bullet$] \textbf{Local Only}: In this method, each client trains its model by feeding local data independently.
    \item[$\bullet$] \textbf{FedAvg}: FedAvg \cite{mcmahan2017communication} applies the averaging method over the weight parameters to obtain a global model. It is a simple but effective way
        to cope with the non-IID scenario. In this method, the parameters are averaged
        in both the reductions and the personalization layers.
    \item[$\bullet$] \textbf{FedProx:} FedProx
        \cite{li2020federated} tackles
        data heterogeneity by adding a proximal term to the loss. We apply the adaptive FedProx loss coefficient in [0.001,0.01,
                0.1,1] based on the fluctuation of the loss.
    \item[$\bullet$]  \textbf{GraphFL:} GraphFL \cite{wang2020graphfl} is a method designed for few-shot learning based on model-agnostic meta-learning (MAML) and addresses the problem of non-IID graph data between clients. We follow the original setting in  \cite{wang2020graphfl} and use 100 nodes for both the support and query set in GraphFL.
    \item[$\bullet$]  \textbf{D-FedGNN:} D-FedGNN \cite{pei2021decentralized} is a distributed federated graph framework based on the weighted communication topology among clients. We follow the setting in  \cite{pei2021decentralized} and use the standard ring network for aggregation.
    \item[$\bullet$]  \textbf{FedGCN:} In FedGCN \cite{yao2022fedgcn}, clients communicate with each other to exchange average information about the node's neighbors. It suffers from severe privacy problems since others know the nodes in a certain client's local dataset. We follow the original setting in  \cite{yao2022fedgcn} and 2 GCN layers for FedGCN.
\end{itemize}
The models are in the same structure with 1 layer MLP as the reduction layers, 2 layers GraphSAGE with activation function as the personalization layers followed by 1 fully connected layer as the classifier.
We choose Relu as the activation function, and Adam as the optimizer.
Besides, the amount of the mashed ego-graphs and the additional communication cost are both strongly influenced by batch size, which is preferred to be 32 based on development test observations \ref{batch_size}.
During the clients' training, nodes will be trained in 5 mini-batches for $5$ epochs each round. In FedEgo, the server will utilize ego-graphs uploaded by clients for training for $5$ epochs each round.
We execute all experiments 4 times and the averaged results are reported
\footnote{Code available at https://github.com/FedEgo/FedEgo}.

\subsection{Overall Perforamance}
\subsubsection{Personalization ability}
The F1 score in the local test given in Table \ref{local_F1} illustrates the personalization ability of each method under severe non-IID scenarios. It is a clear finding that the result in the local test is much higher compared to the global test, primarily because of the same distribution of the testing and training data. Furthermore, FedEgo, FedAvg, FedProx, and D-FedGNN benefit from the collaboration on all datasets and enhance the personalization ability of local models. D-FedGNN performs better than FedAvg since the average merely with local neighbors mitigates the issues of non-IID to some degree.
For FedEgo, there is only a slight performance improvement than FedAvg, which is caused by the label skew scenario and the information from others only compensates for a client's training to a limited extent.
Interestingly, GraphFL performs worse than other FL methods in all cases, owing to the fact that it is designed for few-shot learning and does not extract enough useful information under the severe label distribution skew scenario. Similarly, FedGCN is not suitable for non-IID scenarios and is no match for the local training when it comes to enormous data, as the results on Wiki and CoraFull demonstrate.

\subsubsection{Generalization ability}
As can be seen from the result in Table \ref{global_F1}, the most striking observation emerging from the comparison is that FedEgo consistently outperforms other methods and improves the generalization ability of clients' local models. With the update in reduction and personalization layers, clients gain about 11\%-15\% improvement in performance than local training. The remarkable improvement indicates that FedEgo is able to accomplish the collaboration of clients and tackle the non-IID graph data.

Similar to FedProx, FedAvg provides a boost to some amount, whereas it still falls short of FedEgo. Clients benefit from collaboration in the reduction and personalization layers, as evidenced by the gap between FedAvg and local training. Besides, D-FedGNN performs slightly better than FedAvg due to its unique aggregation.
Compared to FedAvg, FedEgo is a personalized pattern rather than averaging updates in personalization layers. FedEgo significantly exceeds FedAvg and D-FedGNN, implying that a mixture of the local and global models is far superior to a naive average in the severe non-IID scenario.
As for GraphFL and FedGCN, they suffer from the same issue as mentioned above and perform poorly in all cases.

\begin{table*}[t]
    \caption{F1 score for node classification in the global test.}
    \centering
    \begin{tabular}{l|c|c|c|c|c|c|c}
        \toprule
        Dataset  & Local Only & FedAvg & FedProx & GraphFL & D-FedGNN & FedGCN & FedEgo         \\
        \midrule
        Cora     & 0.691      & 0.769  & 0.77    & 0.739   & 0.776    & 0.681  & \textbf{0.794} \\
        Citeseer & 0.634      & 0.701  & 0.701   & 0.649   & 0.712    & 0.655  & \textbf{0.727} \\
        Wiki     & 0.7        & 0.785  & 0.784   & 0.677   & 0.795    & 0.439  & \textbf{0.818} \\
        CoraFull & 0.501      & 0.544  & 0.549   & 0.35    & 0.576    & 0.486  & \textbf{0.653} \\
        \bottomrule
    \end{tabular}
    \label{global_F1}

\end{table*}

\begin{table*}[t]
    \caption{F1 score for node classification in the local test.}
    \centering
    \begin{tabular}{l|c|c|c|c|c|c|c}
        \toprule
        Dataset  & Local Only & FedAvg & FedProx & GraphFL & D-FedGNN & FedGCN & FedEgo         \\
        \midrule
        Cora     & 0.851      & 0.952  & 0.954   & 0.861   & 0.954    & 0.884  & \textbf{0.959} \\
        Citeseer & 0.757      & 0.917  & 0.915   & 0.766   & 0.917    & 0.894  & \textbf{0.920} \\
        Wiki     & 0.874      & 0.922  & 0.924   & 0.789   & 0.915    & 0.819  & \textbf{0.926} \\
        CoraFull & 0.639      & 0.874  & 0.875   & 0.479   & 0.883    & 0.838  & \textbf{0.901} \\
        \bottomrule
    \end{tabular}
    \label{local_F1}

\end{table*}

\subsection{The Effort of Statistical Heterogeneity}
The performance of FedAvg and FedEgo is heavily influenced by the statistical heterogeneity among clients that is controlled by the major node rate.
With a larger major node rate, clients tend to have more nodes in the same classes, leading to a higher statistical heterogeneity. According to this, we vary the major node rate and provide the results in Table \ref{comparison}.
Unsurprisingly, the generalization ability of both FedEgo and FedAvg reduces as the major node rate becomes larger. Perhaps more surprisingly, the increase in major node rate improves the personalization ability to some level. In this case, more nodes in the local dataset will share the same labels and the pattern of interconnections between nodes becomes relatively constant and easy to learn.

\begin{table*}[t]
    \centering
    \begin{minipage}{\textwidth}
        \caption{F1 score on Wiki under different major node rates.
        }
        \label{comparison}
        \centering
        \begin{tabular}{l|c|c|c|c|c|c}
            \toprule
                            & \multicolumn{3}{|c|}{Global Test} & \multicolumn{3}{|c}{Local Test}                                                                     \\
            \cline{2-7}
            Major Node Rate & 0.0                               & 0.3                             & 0.8            & 0.0            & 0.3            & 0.8            \\
            \hline
            FedAvg          & 0.803                             & 0.800                           & 0.785          & \textbf{0.907} & 0.906          & 0.922          \\
            FedEgo          & \textbf{0.823}                    & \textbf{0.826}                  & \textbf{0.818} & 0.903          & \textbf{0.907} & \textbf{0.926} \\
            \bottomrule
        \end{tabular}
    \end{minipage}
\end{table*}

\subsection{Ablation Study}
To verify the effectiveness of Mixup, we undertake a comparison method called \textit{FedEgo w/o Mixup} wherein clients upload the first ego-graph within a batch to the server rather than the mashed ego-graphs. Note that it is not feasible to do so in real scenarios since the transmission of raw ego-graphs may result in the risk of privacy leakage.
We also compare FedEgo with its linear variant, denoted as \textit{FedEgo-Linear}, in order to evaluate the impact of the activation function in the personalization layers. FedEgo-Linear guarantees unbiased estimate of the center nodes' embedding, whereas FedEgo applies activation function to learn more complex patterns in the data.

As the result in Table \ref{comparison_mixup} indicates, Mixup improves both the generalization ability and personalization ability of the model while preserving privacy. The reason is that the graph data among clients are severely non-IID and the virtual samples generated by Mixup can be considered as the clients' exploration of the unknown distribution in the real world dataset, thereby improving the robustness of the model. By comparing FedEgo with FedEgo w/o Mixup, we can also infer that structural information are still highly retained in the mashed ego-graphs after Mixup is applied. The averaging operation in Mixup extracts essential part of the structural information, which facilitates the training in the server and results in better performance of the model.
Further, we observe that the performance marginally decreases without activation function in the personalization layers, which indicates that the model does not necessarily suffer from the bias introduced by the activation function, paricularly in the case of extremely non-IID scenario.

\begin{table*}[t]
    \centering
    \begin{minipage}{\textwidth}
        \caption{Comparison between FedEgo, FedEgo-Linear and FedEgo w/o Mixup.}
        \label{comparison_mixup}
        \centering
        \begin{tabular}{l|c|c|c|c}
            \toprule
                             & \multicolumn{2}{|c|}{Global Test} & \multicolumn{2}{|c}{Local Test}                                   \\
            \cline{2-5}
            Dataset          & Cora                              & Citeseer                        & Cora           & Citeseer       \\
            \hline
            FedEgo w/o Mixup & 0.783                             & 0.717                           & 0.951          & 0.912          \\
            FedEgo-Linear    & 0.793                             & 0.717                           & 0.958          & 0.918          \\
            FedEgo           & \textbf{0.794}                    & \textbf{0.727}                  & \textbf{0.959} & \textbf{0.920} \\
            \bottomrule
        \end{tabular}
    \end{minipage}

\end{table*}

\begin{figure*}[t]
    \centering
    \captionsetup{justification=centering}
    \subfloat[Global test]{
        \includegraphics[width=.235\textwidth]{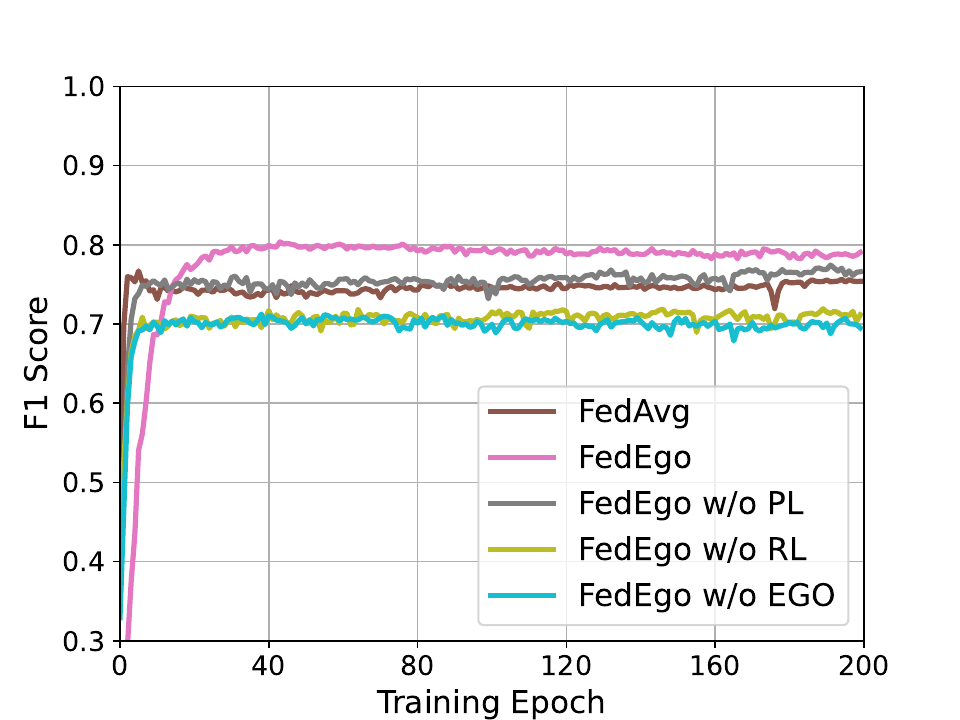}
    }
    \subfloat[Local test]{
        \includegraphics[width=.235\textwidth]{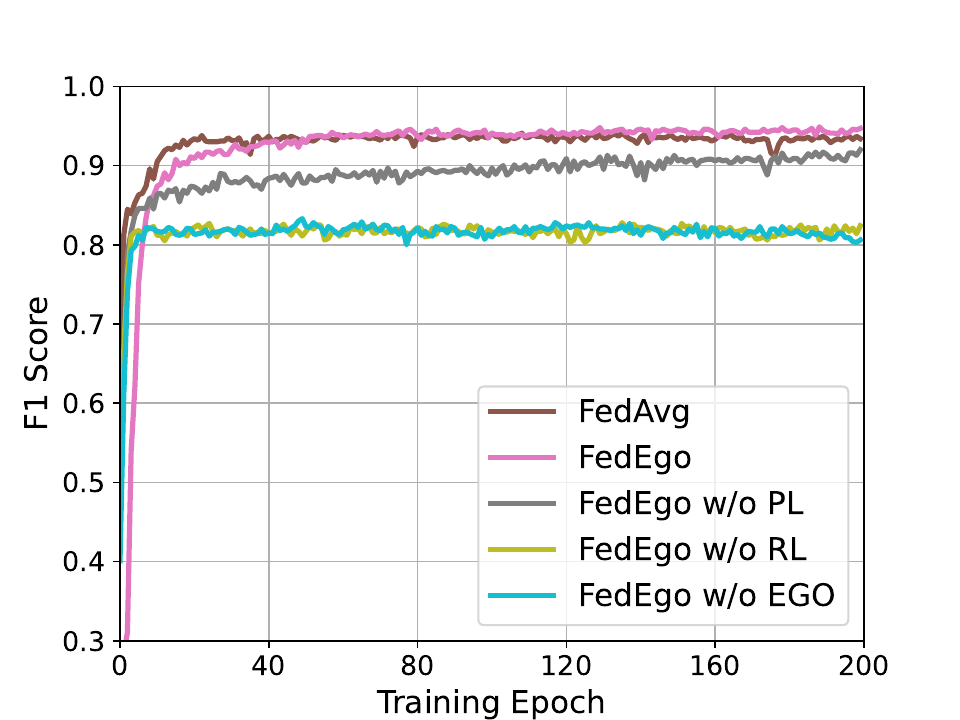}
    }
    %
    \subfloat[Global test]{
        \includegraphics[width=.235\textwidth]{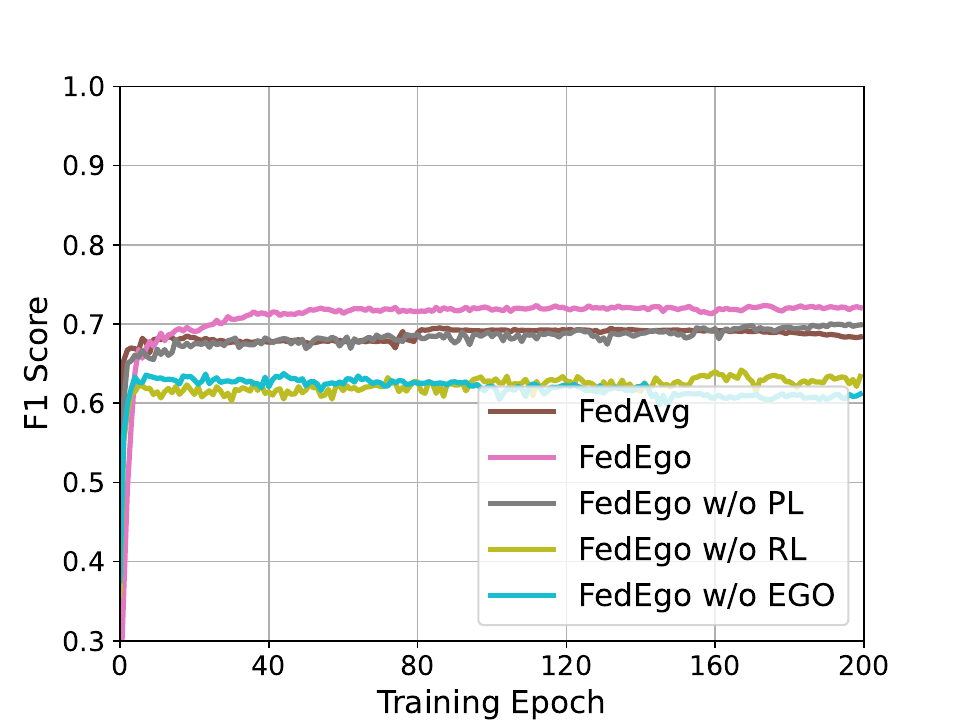}
    }
    \subfloat[Local test]{
        \includegraphics[width=.235\textwidth]{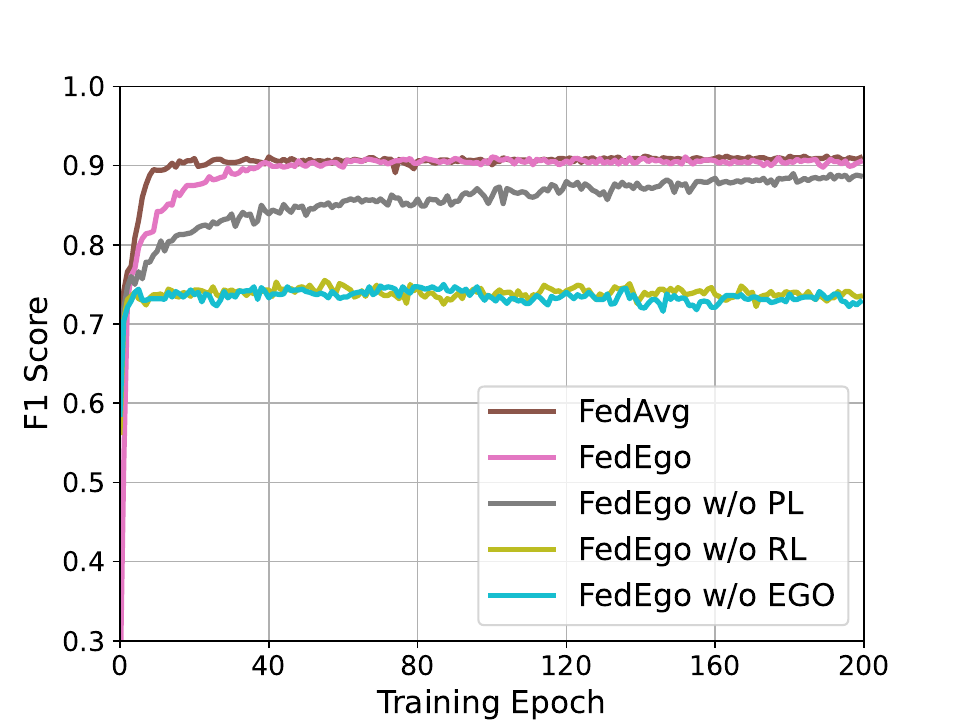}
    }

    \caption{Ablation study on Cora (a,b) and Citeseer (c,d).}
    \label{ablation}

\end{figure*}

To further figure out whether it is worthwhile to build ego-graphs and verify the effectiveness of reduction layers and personalization layers, we compare FedEgo with three ablation methods:
(1) \textit{FedEgo w/o EGO}: In this method, we replace the reduction layers with GNN and the personalizationn layers with MLP.
In this case, graph mining is performed locally and only the reduction embedding need to be uploaded rather than the explict structure of the mashed ego-graphs.
(2) \textit{FedEgo w/o RL}: In this method, clients will only update their personalization layers without performing the averaging updates in reduction layers.
In this case, clients will encrypt the graph data in various ways, making it difficult to train the global model in the server.
(3) \textit{FedEgo w/o PL}: In this method, clients will only update their reduction layers without performing the mixing update in personalization layers.
In this case, clients will not receive the help of the global model but only benefit from averaging update in reduction layers.
Moreover, we also include FedAvg since FedAvg is an averaging version of Fedego w/o PL compared to FedEgo.

The poor performance of FedEgo w/o EGO, as can be seen from Fig. \ref{ablation}, demonstrates the necessity to upload mashed ego-graphs for structural information flow in the federated framework. Without the mashed ego-grpahs, it is rather difficult for the server to capture the structural information solely from the reduction embedding and further provide proper help for clients.
Moreover, it is evidently clear that reduction layers play a more important part in enhancing the performance. Unsurprisingly, the removal of the averaging update in reduction layers hinders clients from encrypting the ego-graphs in the same way, and the mashed ego-graphs uploaded may be even harmful to the global model in early training. Besides, there is still a significant difference between FedEgo w/o PL and FedEgo, demonstrating  the effectiveness of mixing updates in personalization layers.
In the case of FedAvg, it requires the lowest convergence time due to its simple aggregation pattern. The slight gap between FedAvg and FedEgo w/o PL verifies the benefit brought by the averaging update in personalization layers, whereas the performance of FedAvg begins to decline after around 15 epochs before stabilizing in the global test. In the local test, however, the F1 score of FedAvg continues to increase until the end. As a result, FedAvg is inclined to improve clients' personalization ability at the expense of generalization ability.
And finally, the observed difference between FedAvg and FedEgo in the local test is limited while there is a large gap in the global test, as we have discussed before.

\begin{figure*}[t]
    \centering
    \subfloat[$\gamma=0.125$]{
        \includegraphics[width=.23\textwidth]{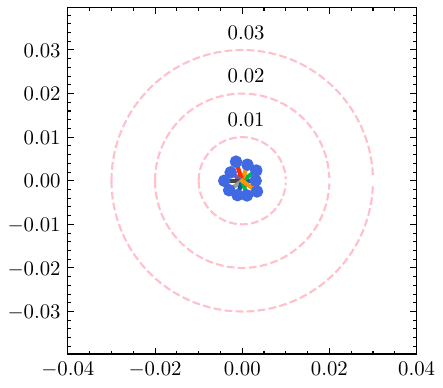}
    }
    \subfloat[$\gamma=0.25$]{
        \includegraphics[width=.23\textwidth]{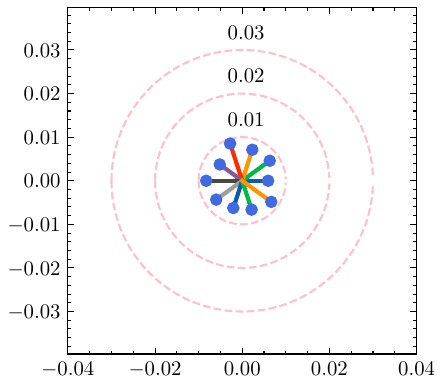}
    }
    \subfloat[$\gamma=0.375$]{
        \includegraphics[width=.23\textwidth]{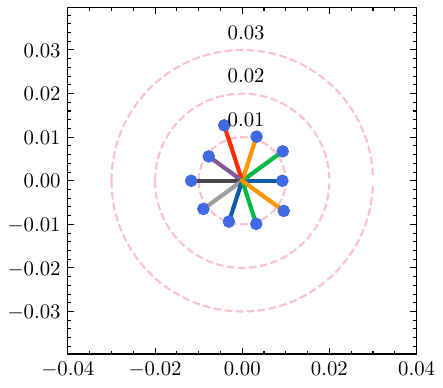}
    }
    \subfloat[$\gamma=0.5$]{
        \includegraphics[width=.23\textwidth]{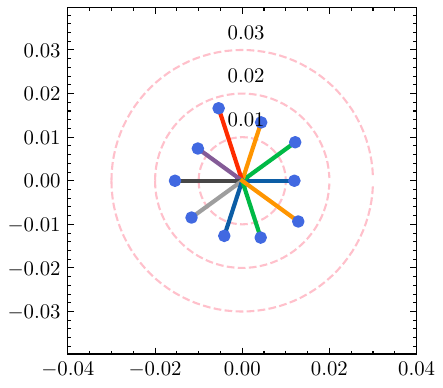}
    }
    \\

    \subfloat[$\gamma=0.625$]{
        \includegraphics[width=.23\textwidth]{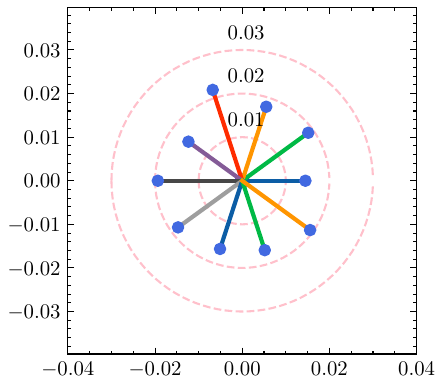}
    }
    \subfloat[$\gamma=0.75$]{
        \includegraphics[width=.23\textwidth]{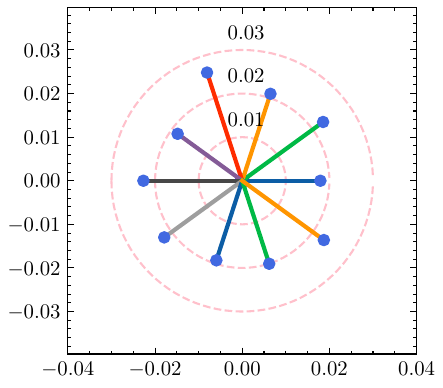}
    }
    \subfloat[$\gamma=0.875$]{
        \includegraphics[width=.23\textwidth]{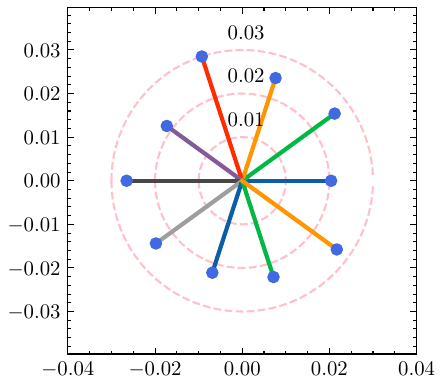}
    }
    \caption{The visualization of weight divergence on Wiki.}
    \label{wd}

\end{figure*}


\subsection{Parameter Study}
To have a better understanding of the hyperparameter $\gamma$, we estimate the weight divergence of each client with the varying $\gamma$. As shown in Fig. \ref{wd}, the global weight is fixed at the center of the circle while scattered points represent the local weight of clients, and the distance to the center indicates the weight divergence. Circles with the same center but different radii are also plotted for reference. As we can see from the result, the weight divergence is affected by $\gamma$ and becomes larger as $\gamma$ increases.
According to Eq. (\ref{lambda}), $\lambda$ is negatively correlated with $\gamma$, and thus the weight divergence increases as $\lambda$ decreases, which is plausible support of Theorem \ref{theorem}.
In addition to the visualization of weight divergence, we provide the F1 score over different values of $\gamma$ in Table \ref{AdaptiveF1}. With a fine-tuned $\gamma$, the adaptive strategy is capable of discovering the best $\lambda$ for each client. Note that the optimal $\gamma$ for the generalization ability and the personalization ability are not always the same due to the statistical heterogeneity. With a relatively large deviation from the global ground-truth, reducing clients' generalization error will likely degrade their performance in the local dataset.

\subsection{Communication Cost Analysis over Batch Size}\label{batch_size}
In FedEgo, clients upload not only the parameters but also the mashed ego-graphs with reduction embedding, and hence whether the additional communication cost is affordable is an significant issue.
A key factor that affects the communication cost and the quality of the mashed ego-graph is the batch size.
With a low batch size, more mashed ego-graphs result in a larger communication cost, and more sensitive information of original data will pass over, yielding the problem of privacy leakage. However, in the other extreme, a large batch size leads to a significant amount of information loss. Based on the analysis above, appropriate batch size is exactly a tradeoff of privacy protection and client collaboration. Therefore, we conduct extensive experiments and evaluate the communication cost in terms of megabytes (MB). The results are provided under various batch sizes in Table \ref{batch}.

The fact that smaller batches lead to higher performance and larger communication cost stands out in the results. The high quality of the uploaded mashed ego-graphs contributes to the improvement, while the communication cost and training over more data make the training time longer. The improvement of reducing batch size decreases as the batch size becomes smaller while the communication overhead, on the other hand, will be doubled due to more batches in the local training.
Furthermore, a smaller batch size fails to meet the need for privacy protection.
Therefore, we believe that selecting 32 as the batch size is a fair tradeoff based on the results.
Additionally, the weight parameters mainly account for the communication cost with such an proper batch size, and it is acceptable to promote the performance at the expense of the additional cost.

\begin{table*}[t]
    \centering
    \begin{minipage}{.7\textwidth}
        \caption{F1 score on CoraFull with different $\gamma$.}
        \label{AdaptiveF1}
        \centering
        \begin{tabular}{l|c|c|c|c|c|c|c}
            \toprule
            $\gamma$    & 0.125 & 0.25           & 0.375 & 0.5   & 0.625 & 0.75           & 0.875 \\
            \hline
            Local Test  & 0.899 & 0.901          & 0.9   & 0.901 & 0.899 & \textbf{0.902} & 0.899 \\
            \hline
            Global Test & 0.652 & \textbf{0.653} & 0.648 & 0.649 & 0.649 & 0.648          & 0.646 \\
            \bottomrule
        \end{tabular}
    \end{minipage}

\end{table*}

\begin{table*}[t]
    \caption{Communication cost analysis for 200 epochs with various batch sizes on Cora and Citeseer.}
    \centering
    \small
    \begin{tabular}{l|c|c|c|c|c|c|c|c|c|c|c|c}
        \toprule
                             & \multicolumn{6}{|c|}{Cora} & \multicolumn{6}{|c}{Citeseer}                                                                                                             \\

        \cline{2-13}
                             & FedAvg                     & \multicolumn{5}{|c|}{FedEgo}  & FedAvg & \multicolumn{5}{|c}{FedEgo}                                                                      \\
        \cline{2-13}
        Batch Size           & 32                         & 8                             & 16     & 32                          & 64    & 128    & 32     & 8      & 16     & 32     & 64    & 128   \\
        \midrule
        Global Test          & 0.769                      & 0.797                         & 0.794  & 0.793                       & 0.792 & 0.764  & 0.701  & 0.724  & 0.724  & 0.725  & 0.72  & 0.711 \\
        Local Test           & 0.952                      & 0.958                         & 0.959  & 0.958                       & 0.955 & 0.945  & 0.917  & 0.92   & 0.919  & 0.918  & 0.918 & 0.917 \\
        Time (mins)          & 10.190                     & 57.674                        & 38.775 & 20.661                      & 15.6  & 10.201 & 11.679 & 90.539 & 41.525 & 32.742 & 13.89 & 6.307 \\
        Ego-graphs Cost (MB) & 0                          & 6071                          & 3055   & 1545                        & 790   & 425    & 0      & 6895   & 3477   & 1769   & 899   & 471   \\
        Parameters Cost (MB) & 6819                       & 2801                          & 2801   & 2801                        & 2801  & 2801   & 11250  & 7234   & 7234   & 7234   & 7234  & 7234  \\
        Total Cost (MB)      & 6819                       & 8872                          & 5856   & 4346                        & 3591  & 3226   & 11250  & 14129  & 10711  & 9003   & 8133  & 7705  \\
        \bottomrule
    \end{tabular}
    \label{batch}

\end{table*}

\subsection{Improvements for each Client}
\begin{figure*}[t]
    \centering
    \captionsetup{justification=centering}
    \begin{minipage}[t]{0.49\textwidth}
        \subfloat[F1 score in the local test]{
            \includegraphics[width=\textwidth]{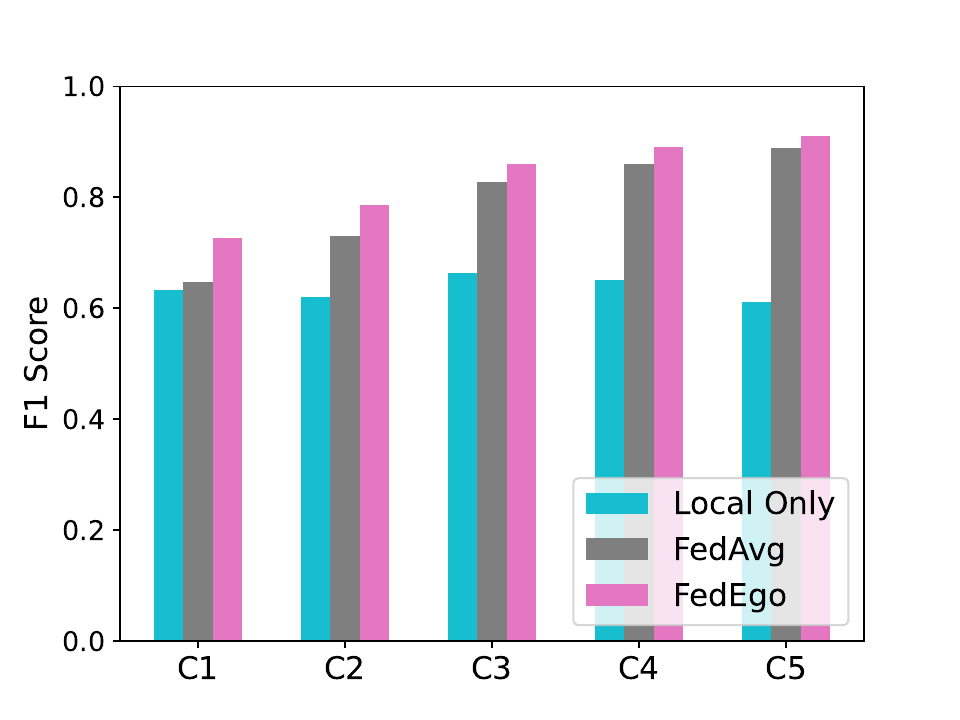}
            \centering
        }
    \end{minipage}
    \begin{minipage}[t]{0.49\textwidth}
        \subfloat[F1 score in the global test]{
            \includegraphics[width=\textwidth]{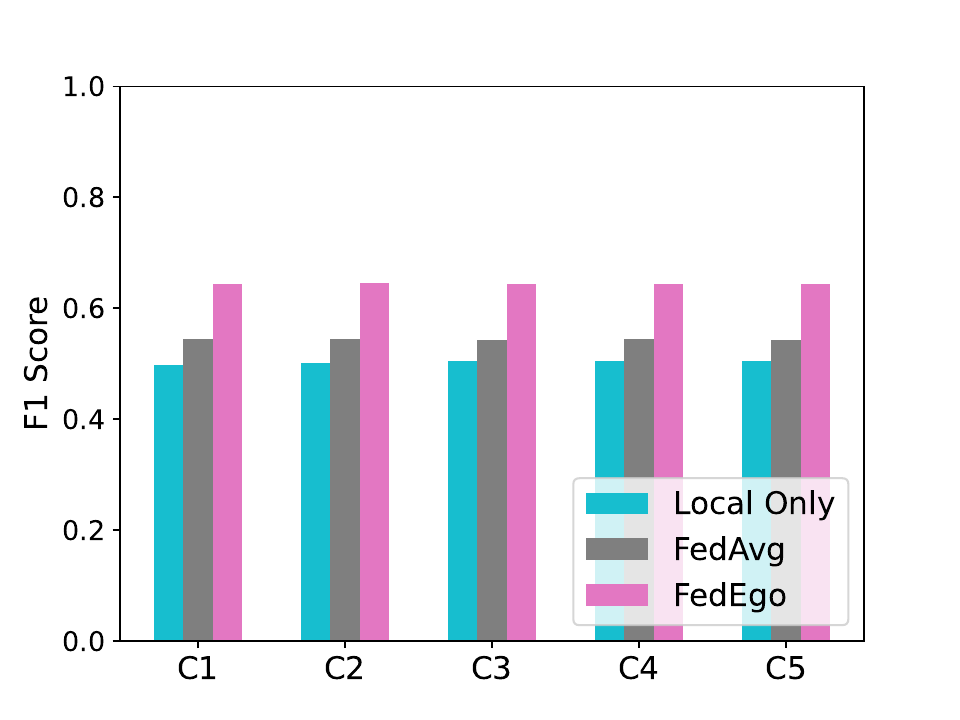}
            \centering
        }
    \end{minipage}
    \caption{F1 score of each client(C1-C5) on CoraFull.}
    \label{each_client}

\end{figure*}

To further explore how FedEgo improves the performance of a specific client, we provide the comparison results of each client in the local and the global test, as shown in Fig. \ref{each_client}. The results indicate that FedEgo enhances the generalization ability and the personalization ability of all the clients. Compared to FedAvg, FedEgo offers more substantial benefits for clients. Moreover, the performance of clients in the global test is relatively stable, even though the graph data are severe non-IID among clients. The stable and significant improvement in performance demonstrates the effect of FedEgo.

\begin{table*}[t]
    \centering
    \begin{minipage}{.7\textwidth}
        \caption{F1 score results in the global test on Wiki with different number of clients.}
        \label{clients_paricipation}
    \end{minipage}
    \begin{minipage}{\textwidth}
        \centering
        \begin{tabular}{l|c|c|c|c|c|c|c}
            \toprule
            Number of Clients & 5               & 10              & 15     & 20     & 25     & 30     & 35     \\
            \hline
            Local Test        & \textbf{0.9261} & 0.9256          & 0.9245 & 0.9255 & 0.9254 & 0.9237 & 0.9247 \\
            \hline
            Global Test       & 0.8179          & \textbf{0.8187} & 0.8175 & 0.8174 & 0.8169 & 0.8158 & 0.8157 \\
            \bottomrule
        \end{tabular}
    \end{minipage}
\end{table*}

\subsection{Effort of Client Participation}
In a real scenario, more clients in participation bring more information and data which may have a potential impact on the statistical heterogeneity. Therefore, we study the influence of client participation on FedEgo in this experiment. In particular, we select 5, 10, 15, 20, 30 and 35 clients to participate in each round on Wiki and the experimental results are reported in Table. \ref{clients_paricipation}.
The personalization ability mainly depends on the client's own dataset and thus slightly drops with an increase in clients due to the introduced bias. As for generalization ability, the global model will have capability to collect more graph data to better fit the real dataset and further develop the generalization ability with more clients participating. This is especially the case when the number of clients is small. However, with a large number of clients, the benefit of introducing new clients diminishes, and the training time increases. Redundant clients might not boost the model performance after there are enough clients is sufficient and the data is no longer in short supply.


\begin{table*}[t]
    \centering
    \begin{minipage}{\textwidth}
        \caption{Analysis of adaptive mixing coefficient.}
        \label{adaptive}
        \centering
        \begin{tabular}{l|c|c|c|c}
            \toprule
                                              & \multicolumn{2}{|c|}{Global Test} & \multicolumn{2}{|c}{Local Test}                                    \\
            \cline{2-5}
            Dataset                           & Cora                              & CoraFull                        & Cora           & CoraFull        \\
            \hline
            FedEgo w/o PL (Fixed $\lambda=0$) & 0.7705                            & 0.6173                          & 0.9345         & 0.8488          \\
            FedEgo (Fixed $\lambda=0.5$)      & 0.7843                            & 0.6437                          & 0.9585         & 0.9005          \\
            FedEgo-Server (Fixed $\lambda=1$) & 0.7921                            & 0.6418                          & 0.9535         & 0.8985          \\
            FedEgo (Adaptive $\lambda$)       & \textbf{0.7939}                   & \textbf{0.6442}                 & \textbf{0.959} & \textbf{0.9012} \\
            \bottomrule
        \end{tabular}
    \end{minipage}

\end{table*}

\begin{figure*}[t]
    \centering
    \captionsetup{justification=centering}
    \begin{minipage}[t]{0.39\textwidth}
        \subfloat[Global ground-truth v.s. local ground-truth]{
            \includegraphics[width=\textwidth]{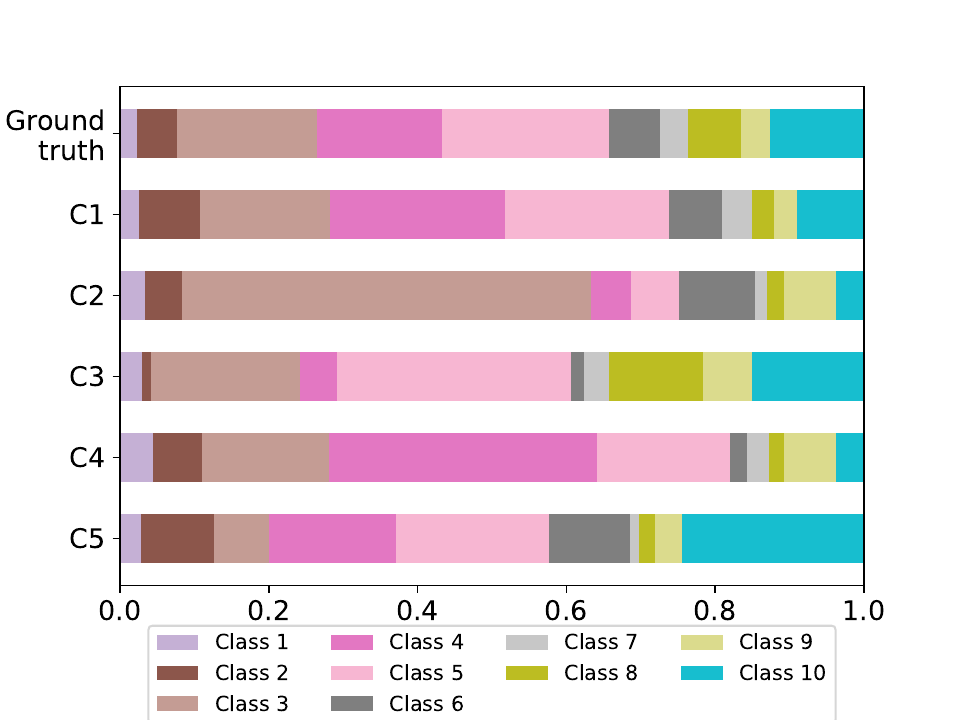}
            \centering
            \label{origin}
        }
    \end{minipage}
    \begin{minipage}[t]{0.39\textwidth}
        \subfloat[Global ground-truth v.s. model predictions]{
            \includegraphics[width=\textwidth]{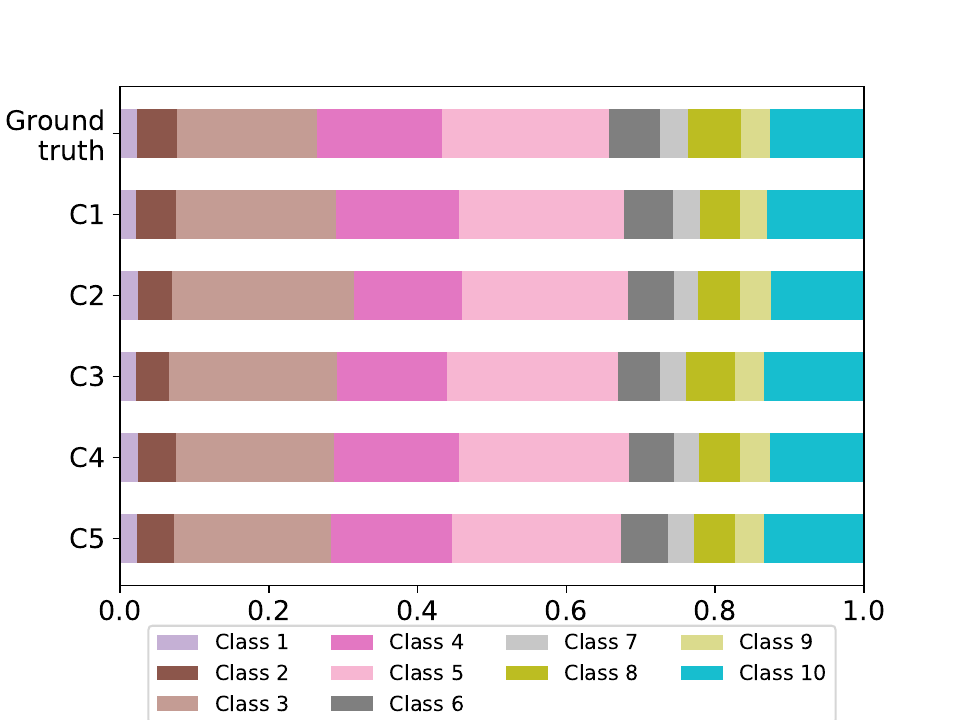}
            \centering
            \label{predict}
        }
    \end{minipage}

    \caption{Label distribution of each client (C1-C5) on Wiki.}
    \label{label_distributio}

\end{figure*}

\subsection{Benefit of Adaptive Mixing Coefficient}
We further conduct additional experiments on Cora and CoraFull to verify
whether FedEgo benefits from the adaptive mixing coefficient $\lambda$.
We compare adaptive $\lambda$ with the following methods:
(1) \textit{FedEgo w/o PL}: As mentioned before, in this method clients would not performing the mixing update in the personalization layers. Thus, it can be considered as a variant when $\lambda$ is fixed to be 0.
(2) \textit{FedEgo (Fixed $\lambda=0.5$)}: In this method, we fixed $\lambda$ to be 0.5 when performing the mixing update in the personalization layers.
(3) \textit{FedEgo-Server}: In this method, clients entirely replace the personalization layers with the weight in the server for graph mining. Thus, it can be considered as a variant when $\lambda$ is fixed to be 1.

As can be seen from Table \ref{adaptive}, adaptive $\lambda$ brings promotion for clients' performance to some extent. In the fixed pattern, FedEgo w/o PL has the worst performance without help of the global model. In the other extreme, FedEgo-Server inferior to FedEgo with adaptive $\lambda$ due to the absence of the personalization. With regard to the compromise strategy, however, the fixed $\lambda=0.5$ may be too small for some clients with larger $EMD$ but too large for those with smaller $EMD$, preventing clients from achieving better performance.
By contrast, the adaptive $\lambda$ allows clients to find their optimal level of participation in federated learning. With adaptive $\lambda$, clients perform better in the local test and achieve better personalization. Meanwhile, appropriate $\lambda$ for each client also improves their F1 score in the global test, which indicates that the mixing coefficient $\lambda$ in personalization layers has an significant influence on the generalization ability of the model.

\subsection{Visulization}
To comprehend how FedEgo improves the performance, we provide the original label distribution and the model prediction of FedEgo. The distribution of 5 local biased and the global testing datasets are shown in Fig. \ref{origin}.
Despite the fact that the graph data are severely non-IID, FedEgo enables clients to effectively extract useful information from their biased local datasets and establish collaboration. As a result, clients are capable of making relatively correct predictions  following the global groud-truth, as shown in Fig. \ref{predict}.

\section{Conclusion}\label{conclusion}

In this paper, we study the personalized federated node classification task on graphs and discuss three main challenges in a realistic setting. The proposed ego-graph-based federated framework FedEgo makes full use of structural information
and tackles non-IID graph data by training a global model in the server. Moreover, clients adapt the global model to its local dataset by mixing the local and global weights. Besides, the Mixup technique is also applied for model robustness and privacy concern. Eventually, FedEgo outperforms baselines significantly with empirical evidence, showing its ability to address the difficult challenges in federated graph learning.
Similar to existing FL methods, future work needs to be done to deal with communication cost for FedEgo, desirably about efficient compression method of ego-graphs and model weight.

\bibliographystyle{ACM-Reference-Format}
\bibliography{FedEgo}


\begin{thebibliography}{27}


\ifx \showCODEN    \undefined \def \showCODEN     #1{\unskip}     \fi
\ifx \showDOI      \undefined \def \showDOI       #1{#1}\fi
\ifx \showISBNx    \undefined \def \showISBNx     #1{\unskip}     \fi
\ifx \showISBNxiii \undefined \def \showISBNxiii  #1{\unskip}     \fi
\ifx \showISSN     \undefined \def \showISSN      #1{\unskip}     \fi
\ifx \showLCCN     \undefined \def \showLCCN      #1{\unskip}     \fi
\ifx \shownote     \undefined \def \shownote      #1{#1}          \fi
\ifx \showarticletitle \undefined \def \showarticletitle #1{#1}   \fi
\ifx \showURL      \undefined \def \showURL       {\relax}        \fi
\providecommand\bibfield[2]{#2}
\providecommand\bibinfo[2]{#2}
\providecommand\natexlab[1]{#1}
\providecommand\showeprint[2][]{arXiv:#2}

\bibitem[Arivazhagan et~al\mbox{.}(2019)]%
        {arivazhagan2019federated}
\bibfield{author}{\bibinfo{person}{Manoj~Ghuhan Arivazhagan},
  \bibinfo{person}{Vinay Aggarwal}, \bibinfo{person}{Aaditya~Kumar Singh},
  {and} \bibinfo{person}{Sunav Choudhary}.} \bibinfo{year}{2019}\natexlab{}.
\newblock \showarticletitle{Federated learning with personalization layers}.
\newblock \bibinfo{journal}{\emph{arXiv preprint arXiv:1912.00818}}
  (\bibinfo{year}{2019}).
\newblock


\bibitem[Bai and Hancock(2016)]%
        {bai2016fast}
\bibfield{author}{\bibinfo{person}{Lu Bai} {and} \bibinfo{person}{Edwin~R
  Hancock}.} \bibinfo{year}{2016}\natexlab{}.
\newblock \showarticletitle{Fast depth-based subgraph kernels for unattributed
  graphs}.
\newblock \bibinfo{journal}{\emph{Pattern Recognition}}  \bibinfo{volume}{50}
  (\bibinfo{year}{2016}), \bibinfo{pages}{233--245}.
\newblock


\bibitem[Bojchevski and G{\"u}nnemann(2017)]%
        {bojchevski2017deep}
\bibfield{author}{\bibinfo{person}{Aleksandar Bojchevski} {and}
  \bibinfo{person}{Stephan G{\"u}nnemann}.} \bibinfo{year}{2017}\natexlab{}.
\newblock \showarticletitle{Deep gaussian embedding of graphs: Unsupervised
  inductive learning via ranking}.
\newblock \bibinfo{journal}{\emph{arXiv preprint arXiv:1707.03815}}
  (\bibinfo{year}{2017}).
\newblock


\bibitem[Cai and Wang(2020)]%
        {cai2020note}
\bibfield{author}{\bibinfo{person}{Chen Cai} {and} \bibinfo{person}{Yusu
  Wang}.} \bibinfo{year}{2020}\natexlab{}.
\newblock \showarticletitle{A note on over-smoothing for graph neural
  networks}.
\newblock \bibinfo{journal}{\emph{arXiv preprint arXiv:2006.13318}}
  (\bibinfo{year}{2020}).
\newblock


\bibitem[Chen et~al\mbox{.}(2021)]%
        {chen2021fedgl}
\bibfield{author}{\bibinfo{person}{Chuan Chen}, \bibinfo{person}{Weibo Hu},
  \bibinfo{person}{Ziyue Xu}, {and} \bibinfo{person}{Zibin Zheng}.}
  \bibinfo{year}{2021}\natexlab{}.
\newblock \showarticletitle{FedGL: federated graph learning framework with
  global self-supervision}.
\newblock \bibinfo{journal}{\emph{arXiv preprint arXiv:2105.03170}}
  (\bibinfo{year}{2021}).
\newblock


\bibitem[Chen et~al\mbox{.}(2018)]%
        {chen2018federated}
\bibfield{author}{\bibinfo{person}{Fei Chen}, \bibinfo{person}{Mi Luo},
  \bibinfo{person}{Zhenhua Dong}, \bibinfo{person}{Zhenguo Li}, {and}
  \bibinfo{person}{Xiuqiang He}.} \bibinfo{year}{2018}\natexlab{}.
\newblock \showarticletitle{Federated meta-learning with fast convergence and
  efficient communication}.
\newblock \bibinfo{journal}{\emph{arXiv preprint arXiv:1802.07876}}
  (\bibinfo{year}{2018}).
\newblock


\bibitem[Collins et~al\mbox{.}(2021)]%
        {collins2021exploiting}
\bibfield{author}{\bibinfo{person}{Liam Collins}, \bibinfo{person}{Hamed
  Hassani}, \bibinfo{person}{Aryan Mokhtari}, {and} \bibinfo{person}{Sanjay
  Shakkottai}.} \bibinfo{year}{2021}\natexlab{}.
\newblock \showarticletitle{Exploiting Shared Representations for Personalized
  Federated Learning}.
\newblock \bibinfo{journal}{\emph{arXiv preprint arXiv:2102.07078}}
  (\bibinfo{year}{2021}).
\newblock


\bibitem[Deng et~al\mbox{.}(2020)]%
        {deng2020adaptive}
\bibfield{author}{\bibinfo{person}{Yuyang Deng},
  \bibinfo{person}{Mohammad~Mahdi Kamani}, {and} \bibinfo{person}{Mehrdad
  Mahdavi}.} \bibinfo{year}{2020}\natexlab{}.
\newblock \showarticletitle{Adaptive personalized federated learning}.
\newblock \bibinfo{journal}{\emph{arXiv preprint arXiv:2003.13461}}
  (\bibinfo{year}{2020}).
\newblock


\bibitem[Hamilton et~al\mbox{.}(2017)]%
        {hamilton2017inductive}
\bibfield{author}{\bibinfo{person}{Will Hamilton}, \bibinfo{person}{Zhitao
  Ying}, {and} \bibinfo{person}{Jure Leskovec}.}
  \bibinfo{year}{2017}\natexlab{}.
\newblock \showarticletitle{Inductive representation learning on large graphs}.
  In \bibinfo{booktitle}{\emph{Advances in Neural Information Processing
  Systems}}. \bibinfo{pages}{1024--1034}.
\newblock


\bibitem[Hanzely and Richt{\'a}rik(2020)]%
        {hanzely2020federated}
\bibfield{author}{\bibinfo{person}{Filip Hanzely} {and} \bibinfo{person}{Peter
  Richt{\'a}rik}.} \bibinfo{year}{2020}\natexlab{}.
\newblock \showarticletitle{Federated learning of a mixture of global and local
  models}.
\newblock \bibinfo{journal}{\emph{arXiv preprint arXiv:2002.05516}}
  (\bibinfo{year}{2020}).
\newblock


\bibitem[He et~al\mbox{.}(2021)]%
        {he2021fedgraphnn}
\bibfield{author}{\bibinfo{person}{Chaoyang He}, \bibinfo{person}{Keshav
  Balasubramanian}, \bibinfo{person}{Emir Ceyani}, \bibinfo{person}{Carl Yang},
  \bibinfo{person}{Han Xie}, \bibinfo{person}{Lichao Sun},
  \bibinfo{person}{Lifang He}, \bibinfo{person}{Liangwei Yang},
  \bibinfo{person}{Philip~S Yu}, \bibinfo{person}{Yu Rong}, {et~al\mbox{.}}}
  \bibinfo{year}{2021}\natexlab{}.
\newblock \showarticletitle{Fedgraphnn: A federated learning system and
  benchmark for graph neural networks}.
\newblock \bibinfo{journal}{\emph{arXiv preprint arXiv:2104.07145}}
  (\bibinfo{year}{2021}).
\newblock


\bibitem[Jiang et~al\mbox{.}(2019)]%
        {jiang2019improving}
\bibfield{author}{\bibinfo{person}{Yihan Jiang}, \bibinfo{person}{Jakub
  Kone{\v{c}}n{\`y}}, \bibinfo{person}{Keith Rush}, {and}
  \bibinfo{person}{Sreeram Kannan}.} \bibinfo{year}{2019}\natexlab{}.
\newblock \showarticletitle{Improving federated learning personalization via
  model agnostic meta learning}.
\newblock \bibinfo{journal}{\emph{arXiv preprint arXiv:1909.12488}}
  (\bibinfo{year}{2019}).
\newblock


\bibitem[Li et~al\mbox{.}(2020)]%
        {li2020federated}
\bibfield{author}{\bibinfo{person}{Tian Li}, \bibinfo{person}{Anit~Kumar Sahu},
  \bibinfo{person}{Manzil Zaheer}, \bibinfo{person}{Maziar Sanjabi},
  \bibinfo{person}{Ameet Talwalkar}, {and} \bibinfo{person}{Virginia Smith}.}
  \bibinfo{year}{2020}\natexlab{}.
\newblock \showarticletitle{Federated optimization in heterogeneous networks}.
\newblock \bibinfo{journal}{\emph{Proceedings of Machine Learning and Systems}}
   \bibinfo{volume}{2} (\bibinfo{year}{2020}), \bibinfo{pages}{429--450}.
\newblock


\bibitem[Mansour et~al\mbox{.}(2020)]%
        {mansour2020three}
\bibfield{author}{\bibinfo{person}{Yishay Mansour}, \bibinfo{person}{Mehryar
  Mohri}, \bibinfo{person}{Jae Ro}, {and} \bibinfo{person}{Ananda~Theertha
  Suresh}.} \bibinfo{year}{2020}\natexlab{}.
\newblock \showarticletitle{Three approaches for personalization with
  applications to federated learning}.
\newblock \bibinfo{journal}{\emph{arXiv preprint arXiv:2002.10619}}
  (\bibinfo{year}{2020}).
\newblock


\bibitem[McMahan et~al\mbox{.}(2017)]%
        {mcmahan2017communication}
\bibfield{author}{\bibinfo{person}{Brendan McMahan}, \bibinfo{person}{Eider
  Moore}, \bibinfo{person}{Daniel Ramage}, \bibinfo{person}{Seth Hampson},
  {and} \bibinfo{person}{Blaise~Aguera y Arcas}.}
  \bibinfo{year}{2017}\natexlab{}.
\newblock \showarticletitle{Communication-efficient learning of deep networks
  from decentralized data}. In \bibinfo{booktitle}{\emph{Artificial
  Intelligence and Statistics}}. PMLR, \bibinfo{pages}{1273--1282}.
\newblock


\bibitem[Mernyei and Cangea(2020)]%
        {mernyei2020wiki}
\bibfield{author}{\bibinfo{person}{P{\'e}ter Mernyei} {and}
  \bibinfo{person}{C{\u{a}}t{\u{a}}lina Cangea}.}
  \bibinfo{year}{2020}\natexlab{}.
\newblock \showarticletitle{Wiki-cs: A wikipedia-based benchmark for graph
  neural networks}.
\newblock \bibinfo{journal}{\emph{arXiv preprint arXiv:2007.02901}}
  (\bibinfo{year}{2020}).
\newblock


\bibitem[Pei et~al\mbox{.}(2021)]%
        {pei2021decentralized}
\bibfield{author}{\bibinfo{person}{Yang Pei}, \bibinfo{person}{Renxin Mao},
  \bibinfo{person}{Yang Liu}, \bibinfo{person}{Chaoran Chen},
  \bibinfo{person}{Shifeng Xu}, \bibinfo{person}{Feng Qiang}, {and}
  \bibinfo{person}{Blue~Elephant Tech}.} \bibinfo{year}{2021}\natexlab{}.
\newblock \showarticletitle{Decentralized Federated Graph Neural Networks}. In
  \bibinfo{booktitle}{\emph{International Workshop on Federated and Transfer
  Learning for Data Sparsity and Confidentiality in Conjunction with IJCAI}}.
\newblock


\bibitem[Sen et~al\mbox{.}(2008)]%
        {sen2008collective}
\bibfield{author}{\bibinfo{person}{Prithviraj Sen}, \bibinfo{person}{Galileo
  Namata}, \bibinfo{person}{Mustafa Bilgic}, \bibinfo{person}{Lise Getoor},
  \bibinfo{person}{Brian Galligher}, {and} \bibinfo{person}{Tina Eliassi-Rad}.}
  \bibinfo{year}{2008}\natexlab{}.
\newblock \showarticletitle{Collective classification in network data}.
\newblock \bibinfo{journal}{\emph{AI magazine}} \bibinfo{volume}{29},
  \bibinfo{number}{3} (\bibinfo{year}{2008}), \bibinfo{pages}{93--93}.
\newblock


\bibitem[Smith et~al\mbox{.}(2017)]%
        {smith2017federated}
\bibfield{author}{\bibinfo{person}{Virginia Smith}, \bibinfo{person}{Chao-Kai
  Chiang}, \bibinfo{person}{Maziar Sanjabi}, {and} \bibinfo{person}{Ameet
  Talwalkar}.} \bibinfo{year}{2017}\natexlab{}.
\newblock \showarticletitle{Federated multi-task learning}.
\newblock \bibinfo{journal}{\emph{arXiv preprint arXiv:1705.10467}}
  (\bibinfo{year}{2017}).
\newblock


\bibitem[Wang et~al\mbox{.}(2020)]%
        {wang2020graphfl}
\bibfield{author}{\bibinfo{person}{Binghui Wang}, \bibinfo{person}{Ang Li},
  \bibinfo{person}{Hai Li}, {and} \bibinfo{person}{Yiran Chen}.}
  \bibinfo{year}{2020}\natexlab{}.
\newblock \showarticletitle{Graphfl: A federated learning framework for
  semi-supervised node classification on graphs}.
\newblock \bibinfo{journal}{\emph{arXiv preprint arXiv:2012.04187}}
  (\bibinfo{year}{2020}).
\newblock


\bibitem[Xie et~al\mbox{.}(2021)]%
        {xie2021federated}
\bibfield{author}{\bibinfo{person}{Han Xie}, \bibinfo{person}{Jing Ma},
  \bibinfo{person}{Li Xiong}, {and} \bibinfo{person}{Carl Yang}.}
  \bibinfo{year}{2021}\natexlab{}.
\newblock \showarticletitle{Federated graph classification over non-iid
  graphs}.
\newblock \bibinfo{journal}{\emph{Advances in Neural Information Processing
  Systems}}  \bibinfo{volume}{34} (\bibinfo{year}{2021}).
\newblock


\bibitem[Yao and Joe-Wong(2022)]%
        {yao2022fedgcn}
\bibfield{author}{\bibinfo{person}{Yuhang Yao} {and} \bibinfo{person}{Carlee
  Joe-Wong}.} \bibinfo{year}{2022}\natexlab{}.
\newblock \showarticletitle{FedGCN: Convergence and Communication Tradeoffs in
  Federated Training of Graph Convolutional Networks}.
\newblock \bibinfo{journal}{\emph{arXiv preprint arXiv:2201.12433}}
  (\bibinfo{year}{2022}).
\newblock


\bibitem[Yoon et~al\mbox{.}(2021)]%
        {yoon2021FedMix}
\bibfield{author}{\bibinfo{person}{Tehrim Yoon}, \bibinfo{person}{Sumin Shin},
  \bibinfo{person}{Sung~Ju Hwang}, {and} \bibinfo{person}{Eunho Yang}.}
  \bibinfo{year}{2021}\natexlab{}.
\newblock \showarticletitle{Fedmix: Approximation of mixup under mean augmented
  federated learning}.
\newblock \bibinfo{journal}{\emph{arXiv preprint arXiv:2107.00233}}
  (\bibinfo{year}{2021}).
\newblock


\bibitem[Zhang et~al\mbox{.}(2017)]%
        {zhang2017mixup}
\bibfield{author}{\bibinfo{person}{Hongyi Zhang}, \bibinfo{person}{Moustapha
  Cisse}, \bibinfo{person}{Yann~N Dauphin}, {and} \bibinfo{person}{David
  Lopez-Paz}.} \bibinfo{year}{2017}\natexlab{}.
\newblock \showarticletitle{mixup: Beyond empirical risk minimization}.
\newblock \bibinfo{journal}{\emph{arXiv preprint arXiv:1710.09412}}
  (\bibinfo{year}{2017}).
\newblock


\bibitem[Zhang et~al\mbox{.}(2021)]%
        {zhang2021subgraph}
\bibfield{author}{\bibinfo{person}{Ke Zhang}, \bibinfo{person}{Carl Yang},
  \bibinfo{person}{Xiaoxiao Li}, \bibinfo{person}{Lichao Sun}, {and}
  \bibinfo{person}{Siu~Ming Yiu}.} \bibinfo{year}{2021}\natexlab{}.
\newblock \showarticletitle{Subgraph federated learning with missing neighbor
  generation}.
\newblock \bibinfo{journal}{\emph{Advances in Neural Information Processing
  Systems}}  \bibinfo{volume}{34} (\bibinfo{year}{2021}).
\newblock


\bibitem[Zhao et~al\mbox{.}(2018)]%
        {zhao2018federated}
\bibfield{author}{\bibinfo{person}{Yue Zhao}, \bibinfo{person}{Meng Li},
  \bibinfo{person}{Liangzhen Lai}, \bibinfo{person}{Naveen Suda},
  \bibinfo{person}{Damon Civin}, {and} \bibinfo{person}{Vikas Chandra}.}
  \bibinfo{year}{2018}\natexlab{}.
\newblock \showarticletitle{Federated learning with non-iid data}.
\newblock \bibinfo{journal}{\emph{arXiv preprint arXiv:1806.00582}}
  (\bibinfo{year}{2018}).
\newblock


\bibitem[Zhu et~al\mbox{.}(2020)]%
        {zhu2020transfer}
\bibfield{author}{\bibinfo{person}{Qi Zhu}, \bibinfo{person}{Yidan Xu},
  \bibinfo{person}{Haonan Wang}, \bibinfo{person}{Chao Zhang},
  \bibinfo{person}{Jiawei Han}, {and} \bibinfo{person}{Carl Yang}.}
  \bibinfo{year}{2020}\natexlab{}.
\newblock \showarticletitle{Transfer learning of graph neural networks with
  ego-graph information maximization}.
\newblock \bibinfo{journal}{\emph{arXiv preprint arXiv:2009.05204}}
  (\bibinfo{year}{2020}).
\newblock


\end{thebibliography}

\appendix

\section{Proof of Theorem \ref{theorem}}\label{prooF1}
In this section, we provide the detailed proof of Theorem \ref{theorem}.
\begin{proof}
    Assume a batch of ego-graphs are in a fixed shape with $K$ hop neighbors and nodes in each layer extends $n$ neighbors.
    For convenience, we sort the nodes based on its layer to assign each node a specific position  and obtain the position sets $Q_{i}$ in the $i$-th layers, i.e., ${Q_0}=\{0\}$, ${Q_1}=\{1, 2,\dots n\}$, ${Q_2}=\{n+1, n+2, \dots n^2\}$ and so on. The node in a specific position  connects to its corresponding neighbors in the next layer. For example, the neighbors of node $1$ in the second layer can be represented as $\{n+1,n+2,\dots 2n\}$. The position of nodes within the same layer can be assigned arbitrarily as long as the special connectivity between layers is maintained. Then the sum of nodes' embedding in a specific layer does not change for different alignment when generating mashed ego-graph. Formally, given a batch of ego-graphs and the mashed ego-graph under an alignment $G$, let $p_i^{(0)}$ be the embedding of node $i$ in the mashed ego-graph, we have
    \begin{equation}\label{sum}
        \sum_{i \in Q_j}p_i^{(0)}\equiv Sum_j, \forall G,
    \end{equation}
    where $Sum_j$ equals the sum of original embedding in the $j$-th layer of the sampled batch ego-graphs.

    Then a $K$ layer of GraphSAGE according to Eq. (\ref{deepLayer}) can be written into the following format.
    \begin{equation}\label{SGC}
        {P^{(K)}}={A}^{K} P^{(0)} W,
    \end{equation}
    where $P$ is the embedding matrix, $A$ denotes the adjacency matrix of the mashed ego-graph and $W$ is a reparameterzed weight matrix by multiplication $W=W^{(1)}W^{(2)}\dots W^{(K)}$. In paricular, $p^{(K)}_0$ denotes the final embedding of the center node.

    For convenience, we further expand $P^{(0)}$ and $A^{K}$ as follows.
    \begin{equation}
        {P^{(0)}}=\begin{bmatrix}
            P^{(0)}_{0} \\
            \vdots      \\
            P^{(0)}_{n^{K}}
        \end{bmatrix}
    \end{equation}

    \begin{equation}
        {A^{K}}=\begin{bmatrix}
            A^{K}_{0,0}     & \cdots & A^{K}_{0,n^{K}}     \\
            \vdots          & \ddots & \vdots              \\
            A^{K}_{n^{K},0} & \cdots & A^{K}_{n^{K},n^{K}}
        \end{bmatrix}
    \end{equation}
    And thus according to Eq. (\ref{SGC}), we have
    \begin{equation}
        {P^{(K)}}=
        \begin{bmatrix}
            A^{K}_{0,0}     & \cdots & A^{K}_{0,n^{K}}     \\
            \vdots          & \ddots & \vdots              \\
            A^{K}_{n^{K},0} & \cdots & A^{K}_{n^{K},n^{K}}
        \end{bmatrix}
        \begin{bmatrix}
            P^{(0)}_{0} \\
            \vdots      \\
            P^{(0)}_{n^{K}}
        \end{bmatrix}
        W
    \end{equation}

    We focus on the center node and compute its final embedding as follows.
    \begin{equation}\label{center}
        {P_0^{(K)}}=
        \begin{bmatrix}
            A^{K}_{0,0} & \cdots & A^{K}_{0,n^{K}}
        \end{bmatrix}
        \begin{bmatrix}
            P^{(0)}_{0} \\
            \vdots      \\
            P^{(0)}_{n^{K}}
        \end{bmatrix}
        W
    \end{equation}

    Due to symmetry of the mashed ego-graph, given a layer $j$, the weight of the edge between node $0$ and each node $i$ in layer $j$ should be the same in the final adjacency matrix $A^{K}$. Thus we have
    \begin{equation}\label{alpha}
        A_{0,i}^{K}\equiv \alpha_j,\forall i \in Q_j, \forall G,
    \end{equation}
    where $\alpha_j$ is a constant that only depends on the $K$ and $n$.

    We continue to combine Eq.(\ref{sum}), (\ref{center}) with (\ref{alpha}) to calculate the final embedding of the center node and complete the proof:
    \begin{equation}\label{complete}
        p_0^{(K)}\equiv (\sum_j^K\alpha_j Sum_j) W, \forall G
    \end{equation}
\end{proof}
\section{Proof of Theorem \ref{theorem2}}\label{proof2}

In this section, we provide the detailed proof of Theorem \ref{theorem2}.
\begin{proof}
    According to the definition of the $\Phi_i^{(T)}$ and $\Phi_g^{(T)}$, we further define the model of epoch $j$ after the $T$-th update as $\Phi_{i,j}^{(T)}$ and $\Phi_{g,j}^{(T)}$.
    For simplicity, the update in personalization layers is defined as the $t$-th step after the $T$-th update. The weight after the update in personalization layers can be written as:
    \begin{equation}
        \left\{
        \begin{aligned}
            \Phi_{i,t}^{(T)} & = \lambda_i\Phi_{g,t-1}^{(T)}+(1-\lambda_i)\Phi_{i,t-1}^{(T)} \\
            \Phi_{g,t}^{(T)} & = \Phi_{g,t-1}^{(T)}
        \end{aligned}
        \right.
    \end{equation}
    And hence we have the change of weight divergence after single update in personalization layers:
    \begin{equation}\label{deeplayer}
        \begin{aligned}
            \|\Phi_{i,t}^{(T)}-\Phi_{g,t}^{(T)}\| =
              & \|\lambda_i\Phi_{g,t-1}^{(T)}+(1-\lambda_i)\Phi_{i,t-1}^{(T)}-\Phi_{g,t-1}^{(T)}\| \\
            = & (1-\lambda_i)\|\Phi_{g,t-1}^{(T)}-\Phi_{g,t-1}^{(T)}\|
        \end{aligned}
    \end{equation}

    Now take the SGD update of the clients and the server into consideration. Given cross-entropy loss defined as Eq. (\ref{loss}), SGD update in the $t-1$-th step perfoms:
    \begin{equation}
        \left\{
        \begin{aligned}
            \Phi_{i,t-1}^{(T)} & =\Phi_{i,t-2}^{(T)}-\eta\nabla_{\Phi} \ell_i\left(\Phi_{i,t-2}^{(T)}\right) \\
            \Phi_{g,t-1}^{(T)} & =\Phi_{g,t-2}^{(T)}-\eta\nabla_{\Phi} \ell_g\left(\Phi_{g,t-2}^{(T)}\right)
        \end{aligned}
        \right.
    \end{equation}
    Therefore, we have:
    \begin{equation}
        \left\{
        \begin{aligned}
            \Phi_{i,t-1}^{(T)} & =\Phi_{i,t-2}^{(T)}-\eta \sum_{c=1}^{C} P_i(y=c) \nabla_{\Phi} \mathbb{E}_{x \mid y=c}[\log f_{c}(x, \Phi_{i,t-2}^{(T)})] \\
            \Phi_{g,t-1}^{(T)} & =\Phi_{g,t-2}^{(T)}-\eta \sum_{c=1}^{C} P_g(y=c) \nabla_{\Phi} \mathbb{E}_{x \mid y=c}[\log f_{c}(x, \Phi_{g,t-2}^{(T)})]
        \end{aligned}
        \right.
    \end{equation}
    We next calculate the weight divergence as follows:
    \begin{equation}
        \begin{aligned}
                   & \|\Phi_{i,t-1}^{(T)}-\Phi_{g,t-1}^{(T)}\|                                                                                                                               \\
            =      & \Bigg\|\Phi_{i,t-2}^{(T)}-\eta \sum_{c=1}^{C} P_i(y=c) \nabla_{\Phi} \mathbb{E}_{x \mid y=c}\left[\log f_{c}\left(x, \Phi_{g,t-2}^{(T)}\right)\right]                   \\
                   & \qquad-\Phi_{g,t-2}^{(T)}+\eta \sum_{c=1}^{C} P_g(y=c) \nabla_{\Phi} \mathbb{E}_{x \mid y=c}\left[\log f_{c}\left(x, \Phi_{g,t-2}^{(T)}\right)\right]
            \Bigg\|                                                                                                                                                                          \\
            {\leq} & \left\|\Phi_{i,t-2}^{(T)}-\Phi_{g,t-2}^{(T)}\right\|
            \\&\qquad+\eta\bigg\|\sum_{c=1}^{C} P_i(y=c)\bigg(\nabla_{{w}} \mathbb{E}_{{x} \mid y=c}\left[\log f_{c}\left({x}, \Phi_{i,t-2}^{(T)}\right)\right]
            \\&\qquad\qquad-\nabla_{{w}} \mathbb{E}_{{x} \mid y=c}\left[\log f_{c}\left({x}, \Phi_{g,t-2}^{(T)}\right)\right]\bigg)         \\
                   & \qquad +\eta \sum_{c=1}^{C}\left(P_i(y=c)-P_g(y=c)\right) \nabla_{{w}} \mathbb{E}_{{x} \mid y=c}\left[\log f_{c}\left({x}, \Phi_{g,t-2}^{(T)}\right)\right]\bigg\|      \\
            {\leq} & \left\|\Phi_{i,t-2}^{(T)}-\Phi_{g,t-2}^{(T)}\right\|                                                                                                                    \\
                   & \qquad +\eta\bigg\|\sum_{c=1}^{C} P_i(y=c)\bigg(\nabla_{{w}} \mathbb{E}_{{x} \mid y=c}\left[\log f_{c}\left({x}, \Phi_{i,t-2}^{(T)}\right)\right]
            \\&\qquad\qquad-\nabla_{{w}} \mathbb{E}_{{x} \mid y=c}\left[\log f_{c}\left({x}, \Phi_{g,t-2}^{(T)}\right)\right]\bigg)\bigg\| \\
                   & \qquad+\eta\left\|\sum_{c=1}^{C}\left(P_i(y=c)-P_g(y=c)\right)\nabla_{{w}} \mathbb{E}_{{x} \mid y=c}\left[\log f_{c}\left({x}, \Phi_{g,t-2}^{(T)}\right)\right]\right\|
        \end{aligned}
    \end{equation}
    The last inequality holds due to absolute value inequality.

    For simplicity, we denotes the gradient of $\mathbb{E}_{{x} \mid y=c}\left[\log f_{c}\left({x}, \Phi_{i,t-2}^{(T)}\right)\right]$ of label $c$ as $g_c$, which only depends on model weight $\Phi$:
    \begin{equation}\label{g_c}
        \begin{aligned}
            g_c(\Phi)=\nabla_{{w}}\mathbb{E}_{{x} \mid y=c}\left[\log f_{c}\left({x}, \Phi\right)\right]
        \end{aligned}
    \end{equation}

    Since $g_c(\Phi)=\nabla_{{w}} \mathbb{E}_{{x} \mid y=c} \log f_{c}({x}, {\Phi})$ is $\L_{{x} \mid y=c}$-Lipschitz, we have
    \begin{equation}\label{absolute}
        \begin{aligned}
            \left\|g_c(\Phi_{i}^{(T)})-g_c(\Phi_{g}^{(T)})\right\|\leq
            L_{{x} \mid y=c}\left\|\Phi_i^{(T)}-\Phi_g^{(T)}\right\|
        \end{aligned}
    \end{equation}

    Then we continue to compute the weight divergence based on Eq.(\ref{g_c}), (\ref{absolute}):
    \begin{equation}
        \begin{aligned}
                 & \left\|\Phi_{i,t-1}^{(T)}-\Phi_{g,t-1}^{(T)}\right\|                                                                                                               \\
            \leq & \left(1+\eta \sum_{c=1}^{C} P_i(y=c) L_{{x} \mid y=c}\right)\left\|\Phi_{i,t-2}^{(T)}-\Phi_{g,t-2}^{(T)}\right\|
            \\&\qquad+\eta\left\|\sum_{c=1}^{C}\left(P_i(y=c)-P_g(y=c)\right)g_c(\Phi_{g,t-2}^{(T)})\right\|                                                                                                         \\
            \leq & \left(1+\eta  L_{max}\sum_{c=1}^{C} P_i(y=c)\right)\left\|\Phi_{i,t-2}^{(T)}-\Phi_{g,t-2}^{(T)}\right\|
            \\&\qquad+\eta g_{m a x}(\Phi_{g,t-2}^{(T)})\sum_{c=1}^{C}\left\|P_i(y=c)-P_g(y=c)\right\| \\
            =    & \left(1+\eta  L_{max}\right)\left\|\Phi_{i,t-2}^{(T)}-\Phi_{g,t-2}^{(T)}\right\|+\eta g_{m a x}(\Phi_{g,t-2}^{(T)})\sum_{c=1}^{C}\left\|P_i(y=c)-P_g(y=c)\right\|,
        \end{aligned}
    \end{equation}
    where $L_{max}=\max _{c=1}^{C} L_{{x} \mid y=c}$ and $ g_{\max }$ is defined as follows.
    \begin{equation}
        g_{\max }\left(\Phi_{g,t-2}^{(T)}\right)=\max _{c=1}^{C}g_c\left(\Phi_{g,t-2}^{(T)}\right)=\max _{c=1}^{C}\left\|\nabla_{{w}} \mathbb{E}_{{x} \mid y=c} \log f_{c}\left({x},\Phi_{g,t-2}^{(T)}\right)\right\|
    \end{equation}

    We further define $a=1+\eta L_{max}$ and we have the following inequality for t-1 steps SGD update by induction:
    \begin{equation}\label{induction}
        \begin{aligned}
                 & \left\|\Phi_{i,t-1}^{(T)}-\Phi_{g,t-1}^{(T)}\right\|                                                                                               \\
            \leq & a\left\|\Phi_{i,t-2}^{(T)}-\Phi_{g,t-2}^{(T)}\right\|+\eta g_{\max }\left(\Phi_{g,t-2}^{(T)}\right) \sum_{c=1}^{C}\left\|P_i(y=c)-P_g(y=i)\right\| \\
            \leq & a^{2}\left\|\Phi_{g,t-3}^{(T)}-\Phi_{i,t-3}^{(T)}\right\|
            \\&\qquad+\eta \sum_{c=1}^{C}\left\|P_i(y=c)-P_g(y=i)\right\|\left(g_{\max }\left(\Phi_{g,t-2}^{(T)}\right)+a g_{\max }\left(\Phi_{g,t-3}^{(T)}\right)\right) \\
            \leq & a^{t-1}\left\|\Phi_{i,0}^{(T)}-\Phi_{g,0}^{(T)}\right\|
            \\&\qquad+\eta \sum_{c=1}^{C}\left\|P_i(y=c)-P_g(y=i)\right\|\left(\sum_{j=0}^{t-2}a^{j} g_{\max }\left(\Phi_{g,t-2-j}^{(T)}\right)\right)                      \\
            =    & a^{t-1}\left\|\Phi_{g,t}^{(T-1)}-\Phi_{g,t}^{(T-1)}\right\|
            \\&\qquad+\eta \sum_{c=1}^{C}\left\|P_i(y=c)-P_g(y=i)\right\|\left(\sum_{j=0}^{t-2}a^{j} g_{\max }\left(\Phi_{g,t-2-j}^{(T)}\right)\right)
        \end{aligned}
    \end{equation}
    Combine Eq.(\ref{deeplayer}) with (\ref{induction}), we bound the weight divergence after $T$-th update and complete the proof:
    \begin{equation}\label{complete}
        \begin{aligned}
                 & \left\|\Phi_{i,t}^{(T)}-\Phi_{g,t}^{(T)}\right\|                                                                                               \\
            =    & (1-\lambda_i)\left\|\Phi_{g,t-1}^{(T)}-\Phi_{g,t-1}^{(T)}\right\|                                                                              \\
            \leq & (1-\lambda_i)a^{t-1}\left\|\Phi_{g,t}^{(T-1)}-\Phi_{g,t}^{(T-1)}\right\|                                                                       \\
                 & +\eta(1-\lambda_i) \sum_{c=1}^{C}\left\|P_i(y=c)-P_g(y=i)\right\|\left(\sum_{j=0}^{t-2}a^{j} g_{\max }\left(\Phi_{g,t-2-j}^{(T)}\right)\right)
        \end{aligned}
    \end{equation}
\end{proof}

\end{document}